\documentclass[10pt,twocolumn,letterpaper]{article}

\usepackage[pagenumbers]{cvpr} 

\usepackage{amsfonts}       %
\usepackage{nicefrac}       %
\usepackage{microtype}      %
\usepackage{arydshln} 
\usepackage{tabularray}
\usepackage{graphicx}
\usepackage{amsmath,amssymb}
\usepackage{amsthm}
\usepackage{multirow}
\usepackage{multicol}
\usepackage{color,xcolor}
\usepackage{booktabs}
\usepackage{threeparttable}
\usepackage{courier}
\usepackage{adjustbox}
\usepackage{supertabular}
\usepackage{stmaryrd}
\usepackage{makecell}
\usepackage{marvosym}
\usepackage{colortbl}
\usepackage{pifont}
\usepackage{appendix}

\usepackage{caption}
\usepackage{times}
\usepackage{epsfig}
\usepackage{graphicx}
\usepackage{verbatim}
\usepackage{bbm}

\usepackage{enumitem,kantlipsum}

\usepackage{fontawesome}
\newcommand \blfootnote[1]{
    \begingroup
        \renewcommand
        \thefootnote{}\footnote{#1}
        \addtocounter{footnote}{-1}
        \vspace{-1ex}
    \endgroup
}

\newcolumntype{g}{>{\columncolor{tbgray}}c}

\newcommand{\Wmat}[0]{{{\bf W}}}

\definecolor{red}{rgb}{0.8, 0.0, 0.0}
\definecolor{green}{rgb}{0.0, 0.5, 0.0}
\definecolor{tbgray}{gray}{.92}

\definecolor{cvprblue}{rgb}{0.21,0.49,0.74}
\usepackage[pagebackref,breaklinks,colorlinks,citecolor=cvprblue]{hyperref}

\def\paperID{16531} 
\def\confName{CVPR}
\def\confYear{2024}

\title{VeCLIP: Improving CLIP Training via Visual-enriched Captions}

\author{$^\text{\faApple}$Zhengfeng Lai$^{1}$\footnotemark[2]\>\,, Haotian Zhang$^{2}$\footnotemark[2]\>\,,  Bowen Zhang$^{2}$, Wentao Wu$^{2}$, Haoping Bai$^{2}$, Aleksei Timofeev$^{2}$, \\
 Xianzhi Du$^{2}$, Zhe Gan$^{2}$, Jiulong Shan$^{2}$, Chen-Nee Chuah$^{1}$, Yinfei Yang$^{2}$\footnotemark[3], Meng Cao$^{2}$ \\
$^{1}$University of California, Davis \quad 
$^{2}$Apple AI/ML\\
\texttt{\{lzhengfeng, chuah\}@ucdavis.edu} \\
\texttt{\{haotian\_zhang2, yinfeiy, mengcao\}@apple.com}
}

\begin{document}
\maketitle
\begin{abstract}
Large-scale web-crawled datasets are fundamental for the success of pre-training vision-language models, such as CLIP. However, the inherent noise and potential irrelevance of web-crawled AltTexts pose challenges in achieving precise image-text alignment. Existing methods utilizing large language models (LLMs) for caption rewriting have shown promise on small, curated datasets like CC3M and CC12M. 
This study introduces a scalable pipeline for noisy caption rewriting. Unlike recent LLM rewriting techniques, we emphasize the incorporation of visual concepts into captions, termed as \textbf{V}isual-\textbf{e}nriched \textbf{Cap}tions (VeCap).
To ensure data diversity, we propose a novel mixed training scheme that optimizes the utilization of AltTexts alongside newly generated VeCap. We showcase the adaptation of this method for training CLIP on large-scale web-crawled datasets, termed VeCLIP. Employing this cost-effective pipeline, we effortlessly scale our dataset up to 300 million samples named VeCap dataset. Our results show significant advantages in image-text alignment and overall model performance. For example, VeCLIP achieves up to  \textbf{+25.2\%} gain in COCO and Flickr30k retrieval tasks under the 12M setting. For data efficiency, VeCLIP achieves  \textbf{+3\%} gain while only using \textbf{14\%} of the data employed in the vanilla CLIP and \textbf{11\%} in ALIGN. 
We also note the VeCap data is complementary with other well curated datasets good for zero-shot classification tasks. When combining VeCap and DFN, our model can achieve strong performance on both of image-text retrieval and zero-shot classification tasks, \emph{e.g.} \textbf{83.1\%} accuracy@1 on ImageNet zero-shot for a H/14 model. We release the pre-trained models at \url{https://github.com/apple/ml-veclip}.
\blfootnote{$^\text{\faApple}$Work done during an internship at Apple. $^\dagger$Equal contribution.}
\blfootnote{$^\ddagger$Corresponding author.}
\end{abstract}
    
\section{Introduction}
\label{sec:intro}

Large-scale vision-language representation learning, exemplified by CLIP~\cite{CLIP}, has gained wide attention due to the transferability of knowledge learned from image-text pairs to diverse downstream tasks such as zero-shot image classification and image-text retrieval~\cite{li2022blip,ALIGN,kwon2022masked}. CLIP training is straightforward via the image-text contrastive loss, but involves a large-scale dataset of 400 million image-text pairs crawled from the Web. 
Consequently, CLIP embeddings lead to consistent improvement across various downstream tasks compared to other vision pre-training methods such as SimCLR~\cite{simclr} and MAE~\cite{MAE}. CLIP achieves success via two scalable paradigms: data and computational resources. First, the massive web-crawled data~\cite{laion_400m,laion5b}
enable the training to be scalable and meet the requirements of data-hungry backbones (\emph{e.g.}, ViT~\cite{vit}). Second, the simple image-text contrastive loss grants favorable scaling properties to the computational resources. 
\begin{figure*}[t]
\begin{center}
\includegraphics[width=1.0\linewidth]{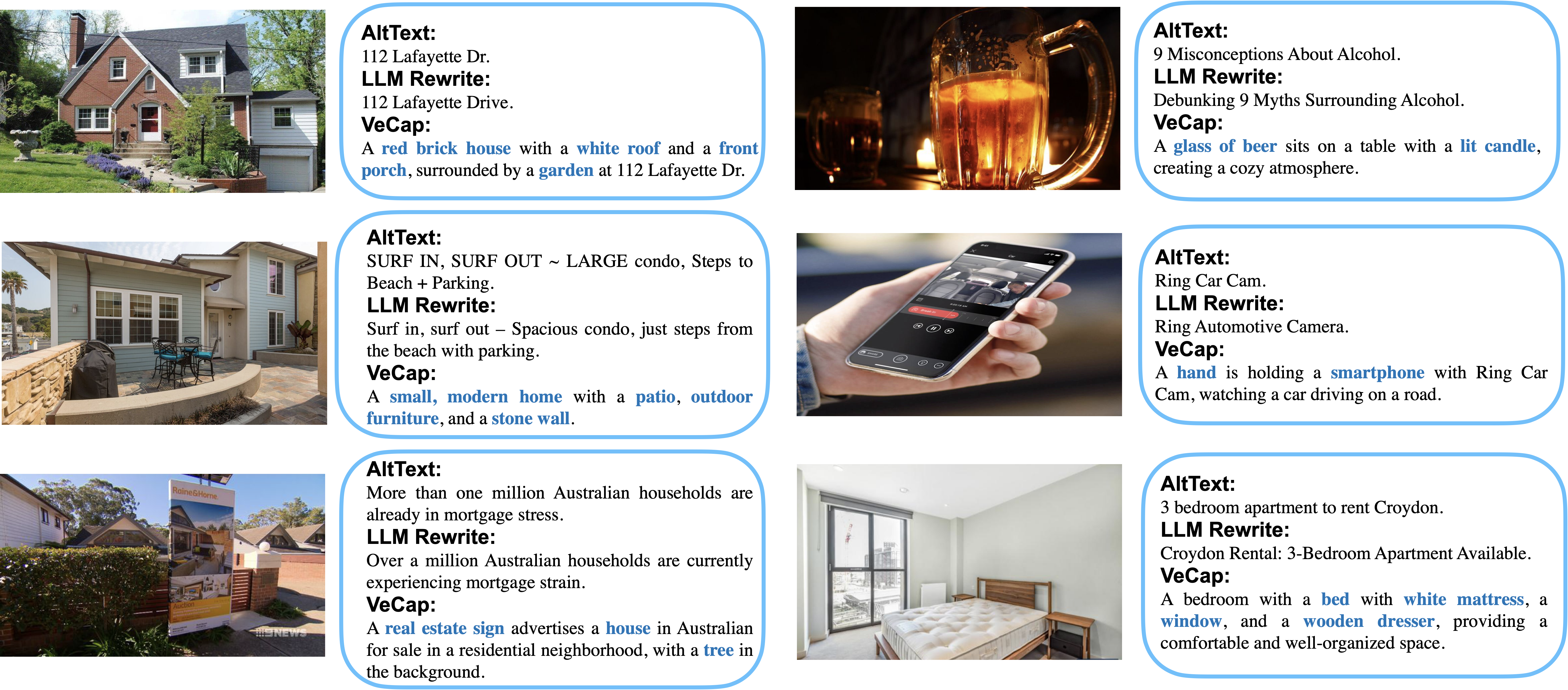}
\end{center}
\vspace{-4mm}
\caption{\footnotesize{\textbf{Noisy web-crawled data and the limitation of LLM rewrite.} AltTexts can be noisy and uninformative; it may not describe all visual objects present in the image. Simple LLM rewrite~\cite{fan2023improveclip} on such raw and noisy captions cannot introduce new image-relevant information. After applying our proposed VeCap, new captions are enriched with more image-specific concepts. We keep all image-text pairs for pre-training rather than filtering out those with noisy AltTexts, as images of rich visual objects still contribute effectively to the training process.  }}
\label{fig:example}
\vspace{-4mm}
\end{figure*}

Despite the availability of large-scale web-crawled data, their quality can be low or noisy. For example, AltTexts suffer from two major issues: 1) they can be noisy, uninformative, or irrelevant to the images; 2) they may not describe all visual contents in the image. For example, as shown in Figure~\ref{fig:example}, in the first image, we observe a house with a white roof and a porch. However, the corresponding caption only describes the address, which proves overly abstract for effective vision-language alignment in training. Our observations demonstrate that caption quality plays a pivotal role in CLIP's performance, as detailed in Table~\ref{table:ab_mixed_captio} and the Appendix (\emph{e.g.}, CC3M vs. our web-crawled 3M). It is worth noting that the captions in CC3M are derived from human annotations, which may require heavy resources when further scaling up. 
This motivates the main open research question addressed in this work: Can we devise a \textbf{scalable and cost-effective pipeline} to improve captions within these noisy datasets at scale (e.g., up to million or billion level)?

One natural direction is to deploy Large Language Models (LLMs) to rephrase captions~\cite{fan2023improveclip}. However, the major limitation of such methods lies in the inability of LLMs to generate and introduce new image-specific details. LLMs can only modify sentence syntax in this scenario. For example, we follow a recent work~\cite{fan2023improveclip} and use LLM to rewrite the raw captions from the Web: as shown in Fig.~\ref{fig:example}, LLM rewrite cannot introduce any new information and thus the new caption remains similar to AltText. In other words, if the original AltTexts are noisy, the benefits brought by LLM rewrite might yield only trivial improvements. In essence, the reliance on high-quality captions within pre-training datasets limits the effectiveness of simple LLM rewrite.
However, sourcing such high-quality datasets like manually curated CC3M and CC12M~\cite{cc12m} remains challenging, and further scaling up to larger datasets becomes both time-consuming and labor-intensive to meet the prerequisites for CLIP pre-training. Therefore, in this work, we focus on building a scalable and cost-effective pipeline tailored to raw and noisy web-crawled data to improve CLIP.

In addition to data quality, the diversity of data significantly impacts VLM pre-training~\cite{nguyen2023captioning,dalle3}. Methods relying on LLM-based rewriting may diminish data variety, given that LLMs tend to apply a uniform style in their sentence rephrasing. Moreover, existing works mainly focus on image augmentations, while texts are disregarded and unaltered during training without augmentation~\cite{fan2023improveclip}. This may also incur overfitting issues as the text encoders repeatedly encounter the same texts in each epoch. Since these techniques have exclusively undergone assessment on meticulously curated datasets like CC3M and CC12M~\cite{cc12m}, their suitability for extensive, uncensored web-crawled data remains uncertain. Consequently, there is a pressing need to build a scalable approach to enhance data quality, diversity, and training methodologies to improve pre-training for VLMs on both model performance and data efficiency.

Concurrently, alongside the evolution of CLIP, there has been substantial progress in the development of instruction fine-tuned LLMs. These models 
and their multimodal extensions
have demonstrated outstanding performance, surpassing human capabilities in various natural language and vision
tasks. 
Inspired by these models, we investigate the potential of utilizing them to improve the noisy captions gathered from the Internet. Specifically, we initially employ LLaVA~\cite{llava}, a Language-Vision Assistant, to leverage visual concepts extracted from the images. 
Given that AltTexts may lack informativeness, our objective is to integrate the newly derived visual concepts into the caption. However, it is worth noting that LLaVA~\cite{llava} fine-tuned its language decoder on its own generated dataset, potentially losing its ability to accommodate comprehensive instructions. Consequently, we further propose to utilize an LLM to refine the sentence by fusing the generated caption from LLaVA and the original AltText. This process aims to maximize image-specific information for optimal vision-language alignment. 
We denote the caption generated from LLM as LLM \textbf{V}isual-\textbf{e}nriched \textbf{Cap}tions~(VeCapap), or \textbf{VeCap} for short. For data variety, we propose VeCLIP and introduce a mixed training scheme, alternating between VeCap and the original AltText. This strategy ensures that the model captures all pertinent information without oversight. We generalize this scalable pipeline to curate five pre-training datasets ranging from small-scale to large-scale up to 300M. 
Overall, our contributions are summarized below:
\begin{itemize}[leftmargin=*]
    \item We present a visual-enriched re-captioning technique for CLIP training. This marks the initial endeavor to leverage visual concepts extracted from images and inject them into the captioning process. 
    \item Our pipeline is cost-effective and capable of processing data at a scale exceeding 300M, named VeCap. Then, we propose VeCLIP with a mixed training scheme that uses VeCap to improve CLIP training on model performance. 
    \item VeCLIP can achieve up to \textbf{25.2\%} improvement over CLIP in retrival tasks. For training data efficiency, \emph{e.g.}, we use only \textbf{5\%} data in training but achieve competitive results in image-text retrieval tasks. 
    \item VeCap data is also complementary with other well curated datasets. A CLIP-H/14 model trained on the combination of VeCap and DFN achieves strong performance on both of image-text retrieval and zero-shot classification tasks, with an impressive \textbf{83.1\%} zero-shot accuracy@1 on ImageNet. 
\end{itemize}


\section{Related Work}
\label{sec:related}

\textbf{Contrastive language-image pre-training.}   CLIP~\cite{CLIP} has shown its effectiveness in acquiring transferable image representations via text supervision after large-scale pre-training. Similar models such as ALIGN~\cite{ALIGN}, Florence~\cite{florence}, BASIC~\cite{pham2021combined} and OpenCLIP~\cite{cherti2023reproducible}
have shown impressive zero-shot image classification and image-text retrieval capabilities. SLIP~\cite{slip} and DeCLIP~\cite{declip} incorporate self-supervised training techniques to improve performance. CoCa~\cite{coca} introduces an additional decoder alongside the contrastive loss. LiT~\cite{lit} proposes to keep a pre-trained image encoder frozen and fine-tune text encoders to improve the zero-shot transferability. Nevertheless, the majority of these subsequent studies incorporate supplementary training inputs and losses, potentially exerting adverse effects on both training efficiency and memory usage.

\textbf{Improving image-text datasets.} 
Given the importance of the pre-training data, many works focus on improving the datasets, such as filtering less informative image-text pairs~\cite{fan2023improveclip,abbas2023semdedup,cao2023less,maini2023t}. However, these methods may disregard a large amount of data even though some images have rich visual concepts. 
An alternative approach is to rewrite the caption to enhance the alignment between texts and images. For example, LaCLIP~\cite{fan2023improveclip} employs LLMs to perform rewriting. Nevertheless, their evaluation was conducted on small-scale and meticulously curated datasets like CC3M and CC12M~\cite{cc12m}, where the initial captions were already of high quality. As shown in Fig.~\ref{fig:example}, the advantage of employing LLM rewrite on noisy web-crawled data is marginal if the AltText is noisy.

\section{Methodology}
\label{sec:methods}
\subsection{Preliminary}

\noindent\textbf{CLIP.} The Contrastive Language-Image Pre-training (CLIP) method 
has shown its effectiveness in training vision models via language supervision. 
Specifically, a batch of $N$ image-text pairs $\{x_I, x_T\}$ is sampled from the massive training data during each training iteration. We apply data augmentations to the images before inputting them into the vision encoder. We denote $f_I$ and $f_T$ as the normalized features extracted by the vision and text encoders, respectively.  We use the contrastive loss to train the model, where the paired images and texts are treated as positive pairs and the remaining as negative samples. 
The training loss iterating over images can be formulated as follows:
\begin{equation}
    L_I = -\sum_{i=1}^N \log  \frac{\exp\big(\text{sim}(f_I(\text{aug}(x_I^i)), f_T(x_T^i))/\tau\big)}{\sum_{k=1}^{N} \exp\big(\text{sim}(f_I(\text{aug}(x_I^i)), f_T(x_T^k))/\tau\big)},
    \label{eq:clip}
\end{equation}
where $(x_I^i, x_T^i)$ is the $i^{th}$ image-text pair in the batch, and $\text{aug}(\cdot)$ refers to image augmentations.  $\text{sim}(\cdot,\cdot)$ is the similarity measurement function. We set $\tau$ as a  learnable temperature parameter that scales the logits in experiments. 
The loss iterating over texts is symmetrical and denoted as $L_T$. Finally, the training loss is $L =( L_I$ + $L_T) / 2$.

\begin{figure*}[t]
\begin{center}
\includegraphics[width=1.0\linewidth]{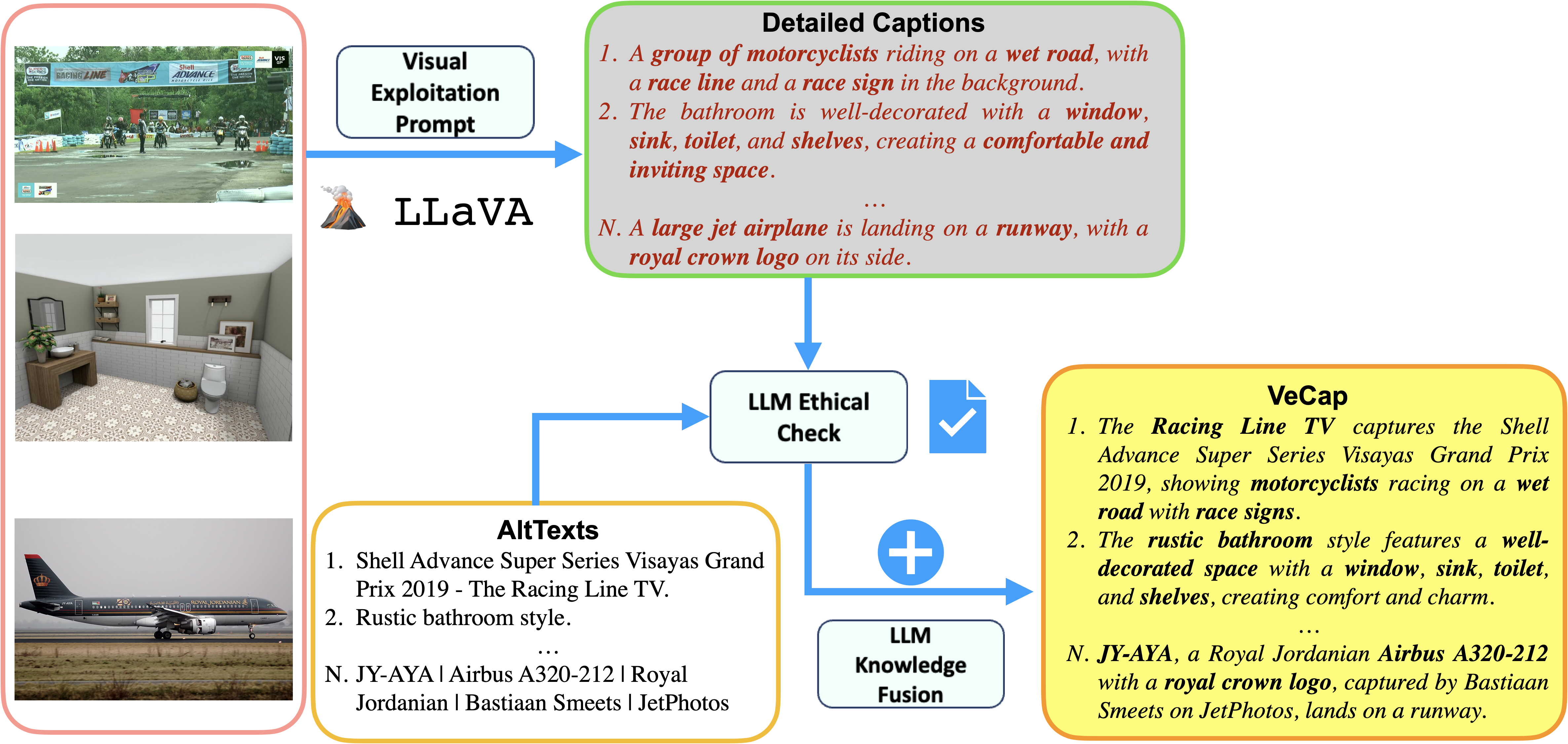}
\end{center}
\caption{\footnotesize{\textbf{An overview of the scalable VeCap recaptioning piepline.} First, we focus on exploiting visual concepts in images via leveraging a multimodal LLM (LLaVA) to describe the image with a designed prompt independent of AltText to generate Visual-enriched Captions (VeC). Second, we leverage an LLM to do ethical check and fuse the concepts from both AltText and VeC to generate the final caption, denoted as VeCap.  }}
\label{fig:pipeline}
\end{figure*}

\subsection{Recaptioning with Visual Concept Exploitation}

Web-crawled captions (AltTexts) can be noisy and uninformative to the images. LaCLIP~\citep{fan2023improveclip} used LLM to rewrite the caption. As shown in Fig.~\ref{fig:example}, this may not be applicable if the captions are noisy as LLM can only reconstruct the sentence but cannot introduce new information without any information provided by the image.
Given the inherent noise in AltTexts, we advocate for the utilization of pre-trained multimodal models to generate augmented captions with richer visual concepts derived from the images. In this subsection, we use LLaVA~\citep{llava} as one example and present a scalable and cost-effective pipeline for scaling up.

\textbf{LLaVA and image captioning for Visual-enriched Captions (VeCap).}
As a multimodal model, LLaVA connects the open-set visual encoder of CLIP~\citep{CLIP} with an LLM, such as LLaMA~\citep{llama}, then fine-tune them on a visual instruction-tuning dataset. LLaVA shows its effectiveness in leveraging the capabilities of pre-trained LLM and vision foundation models. Given an input image $x_I$, we get $f_I$ from CLIP's vision encoder. Then, LLaVA applies a trainable projection matrix $\Wmat$ to convert $f_I$ into language embedding tokens to achieve the image-language alignment. To mitigate the influence of AltText, we have devised AltText-independent prompts tailored for LLaVA, ensuring the full exploitation of visual concepts. We refrain from incorporating AltText information into LLaVA, while acknowledging the potential loss of pre-trained knowledge during fine-tuning of the LLM component on the generated dataset. This trade-off, however, may limit its capacity to comprehend more intricate instructions. Thus, we adopt a straightforward yet potent prompt, \textit{``Describe the image concisely, less than 20 words''}, allowing LLaVA to generate visual concepts directly from the image autonomously.
We denote this captions generated by LLaVA as $x_{Tv}$. Subsequently, the image-text pair is converted as $(x_I, x_{Tv})$.


\subsection{Scalable LLM Rewrite for Concept Fusion}
Given the limited language capacity of LLaVA, we only use LLaVA to extract all possible visual clues. Then, we employ LLMs to refine the caption by fusing both the knowledge from AltText $x_T$ and the novel visual concepts from $x_{Tv}$. 
This step has three main advantages: 1) It ensures the retention of information delineated in AltText, thereby amplifying the informativeness of the caption; 2) It can serve as a form of ``strong augmentation'' in textual data, characterized by a profound restructuring of sentence syntax instead of focusing on word-level modifications used in existing language augmentation techniques~\citep{wei2019eda,sennrich2015improving};
3) It can mitigate the ``hallucination'' issue arising from large vision-language models (e.g., LLaVA) to ensure that the entity described in the ultimate caption is present in the image. 


Generating rewrites for a vast corpus of texts using closed-source models like ChatGPT or Bard is impractical, considering the substantial financial costs and time incurred through API utilization.
Therefore, to facilitate the rewriting tasks on a large-scale dataset, we turn to open-source state-of-the-art LLMs. Due to the license issue, we select Vicuna-1.1~\citep{vicuna}, renowned for its robust performance in text completion tasks, as one example of LLM rewriting in this study. 
We formulate a context input as the following three components. First, we include a sentence designed to apprise the LLM of the task, specifically,  rewriting and fusing two attached sentences. This serves as an initial contextual cue to orient the LLM towards comprehending the overarching objective.
Second, we impose several constraints on the ultimate output. For instance, our goal is to position attributes prior to noun entities, all while refraining from introducing any novel semantic interpretations. Furthermore, it is essential that the sentence refrains from commencing with the phrase ``The image'' and instead directly expounds upon all-encompassed concepts.
Finally, the last part of the context includes two sentences ($x_{v}$ and $x_{Tv}$) that require fusing and rewriting, followed by the separation symbol. This ensures that the LLM is furnished with the specific texts to be fused and rewritten as part of its context input. By integrating these three components, we establish an all-encompassing context that steers the LLM towards proficiently crafting diverse and knowledge-fused text rewrites.

\textbf{Scalable batch-inference process.} 
Employing the crafted context input as a prompt, Vicuna showcases its proficiency in executing text completion and producing rephrased renditions of the associated text samples. However, single-item inference may be time-consuming and not scalable for massive data. Therefore, we conduct this process in a batch-inference process instead of a single-item inference as shown in Fig~\ref{fig:pipeline}: we group our data into batches and implement a batch-inference process to achieve up to 64 times faster  on Nvidia A100. 
Specifically, we use Vicuna-1.1-13B model to generate the final output as  $x_{Tl}$. 
The final prompt is as follows: [\textit{Rephrase the following two sentences into one short sentence while adhering to the provided instructions: Place attributes before noun entities without introducing new meaning. Do not start with ``The image''.  + 1. AltText; 2. model generated caption.}] We denote the caption from LLM as LLM-VeCap, or VeCap for short.

\textbf{Potential ethics of LLM and failure cases processing.  } While upscaling the LLM rewriting process, we identify two scenarios in which LLM encounters difficulties in executing the designated task: 1) \textit{Ethical Concerns}. If the AltText contains content either illegal or violent, LLM may reply, ``I am sorry that I cannot...''; 2) \textit{Length Constraint}. In cases where the AltText exceeds an optimal length, the processing time of the LLM may be significantly prolonged, thus impeding large-scale rewriting.
To address the first scenario, we use the model generation captions as the only input to be rewritten via LLM to form VeCap, thereby preemptively excluding potentially unlawful or aggressive content. In the second scenario, we mitigate this issue by preserving the generated catpion but truncating the AltText to conform to the maximum allowable length, thus we have more visual concepts aligned with the image.




\subsection{VeCLIP: Mixed Training Scheme with Visual-enriched Captions for CLIP}
As LLM rewriting may introduce a consistent style, there could be a decline in data diversity for large-scale pre-training, even if data quality is enhanced. To enhance data diversity, we propose a mixed training scheme to serve as additional text augmentations applied in pre-training:
\begin{equation}
    \text{mix}(x_t) \sim \text{Uniform}([x_{T}, x_{Tl}]).
\end{equation}
Then, the training loss iterating over the images becomes: $L_I = -\sum_{i=1}^N \log  \frac{\exp\big(\text{sim}(f_I(\text{aug}(x_I^i)), f_T({\text{mix}(x_t^i)}))/\tau\big)}{\sum_{k=1}^{N} \exp\big(\text{sim}(f_I(\text{aug}(x_I^i)), f_T({\text{mix}(x_t^k)}))/\tau\big)}.$
The only difference with the original CLIP training is that we alternate the AltTexts with our rephrased sentences, with all other components remaining unaltered. This modification does not incur additional computational complexity or parameter overheads compared to the standard CLIP training process. Through the strategic alternation of AltTexts and our captions, we improve both the quality and diversity of the pre-training dataset without filtering any data points. This approach empowers the model to glean insights from both AltText and VeCap. This simple yet effective strategy elevates the training regimen for CLIP, offering a scalable framework for optimizing other vision-language pre-training efforts utilizing extensive web-crawled data.
\section{Experiments}
\label{sec: results}
\begin{table*}[ht]
    \centering
    \caption{\small{ Results (Recall@$k$) on zero-shot image-to-text and text-to-image retrieval tasks on COCO and Flickr30k. 1.4B-CLIP denotes the in-house CLIP pre-trained on 1.4B web-crawled image-text pairs. We use ViT-B/16 as the vision encoder of CLIP. (*) Denote FLIP uses ViT-L/14. 
}}
    \vspace{-0.2cm}
    \small
    \resizebox{1.0\textwidth}{!}{
    \begin{tabular}{lc|cccccc|cccccc}
  
    \toprule[1.2pt]
    \multirow{3}{*}{\bf Data} & \multirow{3}{*}{\bf Model} & \multicolumn{6}{c|}{\bf COCO}                        & \multicolumn{6}{c}{\bf Flickr30k}    \\                     
    &                   & \multicolumn{3}{c}{ \bf Image-to-Text} & \multicolumn{3}{c|}{\bf Text-to-Image} & \multicolumn{3}{c}{\bf Image-to-Text} & \multicolumn{3}{c}{\bf Text-to-Image} \\
    &                   &  R@1     &   R@5    &  R@10    &   R@1    &     R@5   &  R@10    &  R@1     &   R@5    &  R@10    &   R@1    &     R@5   &  R@10       \\
    
 
    \midrule 
    \small 
    \bf 1.8B & \small ALIGN~\citep{ALIGN} & 58.60 & 83.00 & 89.70 & 45.60 & 69.80 & 78.60 & 88.60 & \bf 98.70 & \bf 99.70 & 75.70 & \bf 93.80 & \bf 96.80 \\
    \bf 400M & \small FLIP*~\citep{flip} & 60.20 & 82.60 & 89.90 & 44.20 & 69.20 & 78.40 & 89.10 & 98.50 & 99.60 & 75.40 & 92.50 & 95.90 \\
    \midrule
    \bf 400M & \small  OpenAI CLIP & 53.76 & 77.92 & 85.53 & 33.09 & 58.42 & 68.90 & 88.00 & \bf 98.70 & 99.40 & 68.70 & 90.60 & 95.20 \\
    \bf 1.4B & \small In-house CLIP & 61.38	& 82.80 &	89.78	&44.48	&69.19 &	78.28
    & 87.60	& 97.90	& 98.80 &	71.70	& 91.30	& 95.24 \\
    \midrule
    
    \multirow{2}{*}{\bf 3M} & \small CLIP & 5.46	&15.34	&22.42	&3.28	&10.44	& 15.96 & 12.20	& 27.80	& 37.50	& 6.36	&19.16	&27.58  \\
    &  \small VeCLIP & \bf 22.30	& \bf 45.00 & \bf	56.16 & \bf	13.01 &	\bf 31.61	& \bf 42.42 & \bf 40.60	& \bf 67.30	& \bf 76.70	& \bf 27.58 & 	\bf 52.44 & 	\bf 63.20 \\
    \rowcolor{tbgray}  \multicolumn{2}{c|}{\bf Performance Gain} & \bf \textcolor{green}{+16.84}	& \bf \textcolor{green}{+29.66}	& \bf \textcolor{green}{+33.74}	& \bf \textcolor{green}{+9.73} 	& \bf \textcolor{green}{+21.17}	& \bf \textcolor{green}{+26.46} & \bf \textcolor{green}{+28.40}	& \bf \textcolor{green}{+39.50}	& \bf \textcolor{green}{+39.20}	& \bf \textcolor{green}{+21.22} & \bf \textcolor{green}{+33.28}	& \bf \textcolor{green}{+35.62} \\

    \midrule 
    \multirow{2}{*}{\bf 12M} & \small CLIP & 24.52	& 48.28	& 59.82 &	14.28	& 34.52 &	46.29 & 44.70	& 71.80 &	80.40 &	29.06	& 58.62 &	70.00 \\
    &  \small VeCLIP & \bf 47.78	& \bf 72.54	& \bf 81.56	& \bf 31.62	& \bf 57.19	& \bf 68.47 & \bf 73.90	& \bf 92.30	& \bf 95.90	& \bf 55.68	& \bf 80.78	& \bf 87.64 \\
    \rowcolor{tbgray}  \multicolumn{2}{c|}{\bf Performance Gain} & \bf \textcolor{green}{+23.26}	& \bf \textcolor{green}{+24.26}	& \bf \textcolor{green}{+21.74}	& \bf \textcolor{green}{+17.34} 	& \bf \textcolor{green}{+22.67}	& \bf \textcolor{green}{+22.18} & \bf \textcolor{green}{+29.20}	& \bf \textcolor{green}{+20.50}	& \bf \textcolor{green}{+15.50}	& \bf \textcolor{green}{+26.62} & \bf \textcolor{green}{+22.16}	& \bf \textcolor{green}{+17.64} \\
    \midrule 
    \multirow{2}{*}{\bf 100M} & \small CLIP & 47.24	&72.34	&81.56&	30.61	& 56.49	& 67.91 & 74.40	& 93.20	& 96.70 &	57.16 & 	88.12 &	88.98\\
    &  \small VeCLIP &\bf   64.82	& \bf 85.56& \bf 91.98	& \bf 46.12	& \bf 71.19	& \bf 80.23 & \bf  89.30	& \bf   97.70	& \bf  99.20 &	\bf  73.10	& \bf  89.12	& \bf 93.14 \\
    \rowcolor{tbgray}  \multicolumn{2}{c|}{\bf Performance Gain} & \bf \textcolor{green}{+17.58}	& \bf \textcolor{green}{+13.22}	& \bf \textcolor{green}{+10.42}	& \bf \textcolor{green}{+15.51} & \bf \textcolor{green}{+14.70}	& \bf \textcolor{green}{+12.32} & \bf \textcolor{green}{+14.90}	& \bf \textcolor{green}{+4.50}	& \bf \textcolor{green}{+2.50}	& \bf \textcolor{green}{+15.94} & \bf \textcolor{green}{+1.00}	& \bf \textcolor{green}{+4.16} \\

    \midrule 
    \multirow{2}{*}{\bf 200M} & \small CLIP &  52.20	& 76.22	& 85.04	& 34.97	& 60.42	& 71.08 & 80.90	& 94.90	& 97.60	& 63.26 &	86.58	& 92.26 \\
    &  \small VeCLIP & \bf 67.20	& \bf 87.28 &  \bf	92.70	& \bf 48.40	& \bf 73.26 & \bf	81.79 &  \bf 91.10 &	 \bf 98.50 & \bf	99.70	& \bf 76.32	& \bf 93.50	& \bf 96.40 \\
    \rowcolor{tbgray}  \multicolumn{2}{c|}{\bf Performance Gain} & \bf \textcolor{green}{+15.00}	& \bf \textcolor{green}{+11.06}	& \bf \textcolor{green}{+7.66}	& \bf \textcolor{green}{+13.43} & \bf \textcolor{green}{+12.84}	& \bf \textcolor{green}{+10.71} & \bf \textcolor{green}{+10.20}	& \bf \textcolor{green}{+3.60}	& \bf \textcolor{green}{+2.10}	& \bf \textcolor{green}{+13.06} & \bf \textcolor{green}{+6.92}	& \bf \textcolor{green}{+4.14} \\

    \midrule
    \multirow{2}{*}{\bf 300M} & \small CLIP &  54.24	& 78.14	& 86.48	& 36.98	& 62.32	& 72.70 & 81.30	& 95.80	& 97.80	& 65.80 &	88.28	& 93.16 \\
    &  \small VeCLIP & \bf 67.80	& \bf 87.94 &  \bf	92.84	& \bf 48.91	& \bf 73.54 & \bf	82.11 &  \bf 91.20 &	 \bf 99.10 & \bf	99.80	& \bf 76.30	& \bf 93.00	& \bf 96.44 \\
    \rowcolor{tbgray}  \multicolumn{2}{c|}{\bf Performance Gain} & \bf \textcolor{green}{+13.56}	& \bf \textcolor{green}{+9.80}	& \bf \textcolor{green}{+6.36}	& \bf \textcolor{green}{+11.93} & \bf \textcolor{green}{+11.22}	& \bf \textcolor{green}{+9.41} & \bf \textcolor{green}{+9.90}	& \bf \textcolor{green}{+3.30}	& \bf \textcolor{green}{+2.00}	& \bf \textcolor{green}{+10.50} & \bf \textcolor{green}{+4.72}	& \bf \textcolor{green}{+3.28} \\   

    \bottomrule[1.2pt]
    \end{tabular}
    }
    \label{tab:retrieval_appendix}
\end{table*}
\subsection{Pre-training Datasets and Downstream Tasks}
\textbf{Pre-training datasets and training setup.}  We conduct pre-training experiments on four scales of our datasets (named VeCap) to show the efficiency and scalability of our method. Specifically, we set 3M as small scale, 12M as medium scale, and 100M+ as large scale. We use ViT-B/16~\cite{vit} as the vision encoder of CLIP training. Our batch size is 8,192 for small/medium scales (3M/12M), and 32,768 for large scales (100M+). For efficiency purposes, we employ \textbf{JAX}~\cite{jax2018github} and train models on 64 TPUs for the 3M/12M settings, whereas we utilize 512 TPUs for the 100M/200M pre-training configurations. 
All models are trained with the AXLearn framework.~\footnote{\href{https://github.com/apple/axlearn}{https://github.com/apple/axlearn}}
More details can be found in the Appendix A. 
To show its generalizability and effectiveness, we also evaluate it on well-curated CC3M/CC12M besides our crawled noisy WIT data, as shown in our ablation studies and Appendix C.2. 
We evaluate all pre-trained models on the following three tasks. 

\textbf{Zero-shot image classification. }
We evaluate all the models on ImageNet~\cite{imagenet}, ImageNetV2~\cite{imagenet_v2}, and VTAB~\cite{vtab}. We select 9 tasks  (6 from natural sets and 3 from specialized sets)  that are suitable for zero-shot classification tasks such as Flowers102~\cite{flowers102} and Caltech-101~\cite{caltech101} as zero-shot classification tasks. 
We list the details in the Appendix. 

\textbf{Zero-shot image-text retrieval. } We evaluate the pre-trained models on COCO~\cite{mscoco} and Flickr30k~\cite{flickr30k} cross-modal retrieval tasks: Image-to-Text (denoted as I2T) and Text-to-Image (T2I) retrieval. For Flickr30k, we evaluate them on the standard 1K test set.  We report the results in terms of Recall@$k$  as R@1, R@5, and R@10. 

\textbf{Zero-shot image-to-image retrieval.} We select GPR1200~\cite{gpr1200} for image-to-image retrieval. 
GPR1200~\cite{gpr1200} serves as a general-purpose benchmark for content-based image retrieval, encompassing subsets drawn from six different domains. It includes 1200 categories (10 images per category). Following~\cite{gpr1200}, we do not split images as query and index sets for evaluation. 
Instead, we perform retrieval of the nearest neighbor for each image and utilize the remaining images as the index set. We report the mean Average Precision (mAP).



\subsection{Results on Retrieval Tasks}
\textbf{I2T and T2I retrieval.} We summarize the main results in Table~\ref{tab:retrieval_appendix}. 
We show consistent improvements across  Recall@$k$ metrics in both COCO and Flickr30k datasets for both I2T and T2I retrieval tasks. Specifically, for small and medium scales (3M/12M), we attain an improvement of +16.84\%/+23.26\% in Recall@1 for COCO image-to-text retrieval, respectively. Notably, the strides made in Flickr30k are particularly noteworthy, with a remarkable +28.40\%/+29.20\% improvement in Recall@1. Subsequently, we scale our approach to 100M and 200M, where we observe sustained and substantial improvements. Notably, we achieve a noteworthy +17.58\%/+15.00\% enhancement in COCO image-to-text retrieval performance using 100M and 200M, respectively. Furthermore, we observe a diminishing improvement margin as we scale up the dataset. Initially, we achieve a substantial 28.40\% improvement in image-to-text retrieval for Flickr30k with the 3M dataset, which subsequently decreases to 10.20\% when employing the 200M dataset. 
These findings show the advantages of our proposed pipeline for enhancing CLIP pre-training.
By demonstrating its scalability from 3M to 300M, we provide compelling evidence of its applicability in real-world scenarios, particularly for training CLIP from scratch using WIT datasets.


\begin{table*}[t]
\begin{tabular}{cc}

\begin{minipage}[t]{0.55\linewidth}
\vspace{0pt}
\centering

\caption{\small{Image-to-image retrieval results (mAP) on 6-domain GPR1200~\citep{gpr1200}. 
}}
\vspace{-0.2cm}
\label{tab:gpr1200}   
\small
\resizebox{1.0\textwidth}{!}{
\begin{tabular}{cc|cccccc|g}
\toprule[1.2pt]
\multirow{2}{*}{\bf Data} & \multirow{2}{*}{\bf Model} & \multicolumn{6}{c|}{\bf Domain Name}  & \\

& & Land & Faces & iNat   & INST & Sketch & SOP & \bf  All \\
\midrule 
\small 
\multirow{2}{*}{\bf 3M} & \small CLIP & 57.98 & 20.76& 17.61 & 31.14	& 18.23	& 74.29 & 36.67 \\
&  \small VeCLIP  & \bf 66.55	& \bf 23.51	&  \bf20.43 & \bf 38.63	& \bf 24.59	& \bf 77.65 & \bf 41.89\\

\midrule
\multirow{2}{*}{\bf 12M} & \small CLIP & 74.47 & 30.65 & 23.60 & 52.15	& 30.68	& 84.25 & 49.30 \\
&  \small VeCLIP  & \bf 79.30	& \bf 31.72 & \bf 25.53 & \bf 56.65	& \bf 41.42	& \bf 84.69 & \bf 53.22\\

\midrule
\multirow{2}{*}{\bf 100M} & \small CLIP & \bf 85.64 & \bf 51.68 & 29.66 & 68.19	& 42.45	& 90.38 & 61.33 \\
&  \small VeCLIP  & 85.59 	&  42.83 & \bf 30.72 & \bf 71.96	& \bf 52.59	& \bf 90.54 & \bf 62.37\\

\midrule
\multirow{2}{*}{\bf 200M} & \small CLIP & \bf 86.96 &\bf 56.54 & 30.95 & 71.51	& 46.03	& 90.95 & 63.83 \\
&  \small VeCLIP  & 86.40 	&  48.48 & \bf 31.72 & \bf 73.74	& \bf 56.52	& \bf 91.16 & \bf 65.67 \\

\bottomrule[1.2pt]
\end{tabular}
}

\vspace{0.3cm} 

\centering
\caption{\small{Zero-shot classification results (Top-$k$ Accuracy) on ImageNet and ImageNetV2~\citep{imagenet_v2}.
}}
\vspace{-0.2cm}
\label{tab:imagenet}   
\small
\resizebox{1.0\textwidth}{!}{
\begin{tabular}{cc|ccc|ccc}
\toprule[1.2pt]
\multirow{2}{*}{\bf Data } & \multirow{2}{*}{\bf Model} & \multicolumn{3}{c|}{ \bf ImageNet} & \multicolumn{3}{c}{ \bf ImageNetV2} \\
& & Top-1 & Top-5 & Top-10   & Top-1 & Top-5 & Top-10  \\
\midrule 
\small 
\multirow{2}{*}{\bf 3M} & \small CLIP & 5.46 & 21.05 & 28.70 & 7.09	& 18.52	& 25.83 \\
&  \small VeCLIP  & \bf 15.98	& \bf 34.11	& \bf 43.23 & \bf 13.51	& \bf 30.03	& \bf 38.93 \\

\midrule
\multirow{2}{*}{\bf 12M} & \small CLIP & 31.60	& 58.80	& 69.49 & 27.03 &	52.68	& 63.37 \\
&  \small VeCLIP  & \bf 38.11	& \bf 66.74	& \bf 76.36 & \bf 32.53	& \bf 60.16 &	\bf 70.50 \\

\midrule
\multirow{2}{*}{\bf 100M} & \small CLIP & 58.64	& 85.82 &	91.79 & 50.96	& 79.77	& 86.91 \\
&  \small VeCLIP  & \bf 60.77	& \bf 87.77	& \bf 93.16 & \bf 54.17	& \bf 82.51 &	\bf 89.24 \\

\midrule
\multirow{2}{*}{\bf 200M} & \small CLIP & 63.72	& 89.26 & 94.11 & 56.84 & 83.50 & 89.79 \\
&  \small VeCLIP  & \bf 64.62	& \bf 90.27	& \bf 94.90 & \bf 57.67	& \bf 85.24 &	\bf 91.62 \\

\bottomrule[1.2pt]
\end{tabular}
}

\end{minipage}

&

\begin{minipage}[t]{0.38\linewidth}
\vspace{0pt}

\includegraphics[width=.99\linewidth, height=0.75\linewidth]{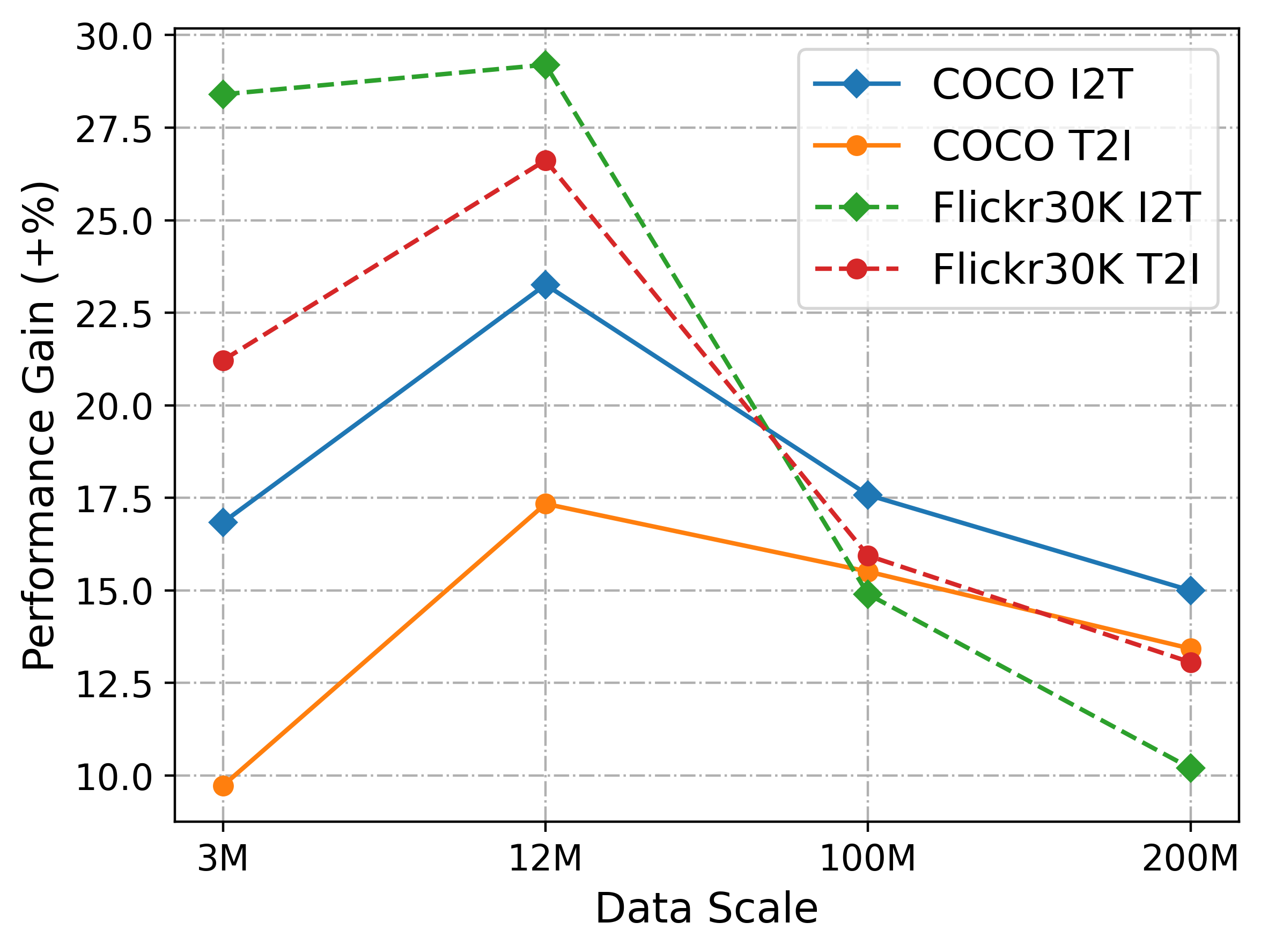}
\vspace{-0.5cm}

\includegraphics[width=.99\linewidth, height=0.75\linewidth]{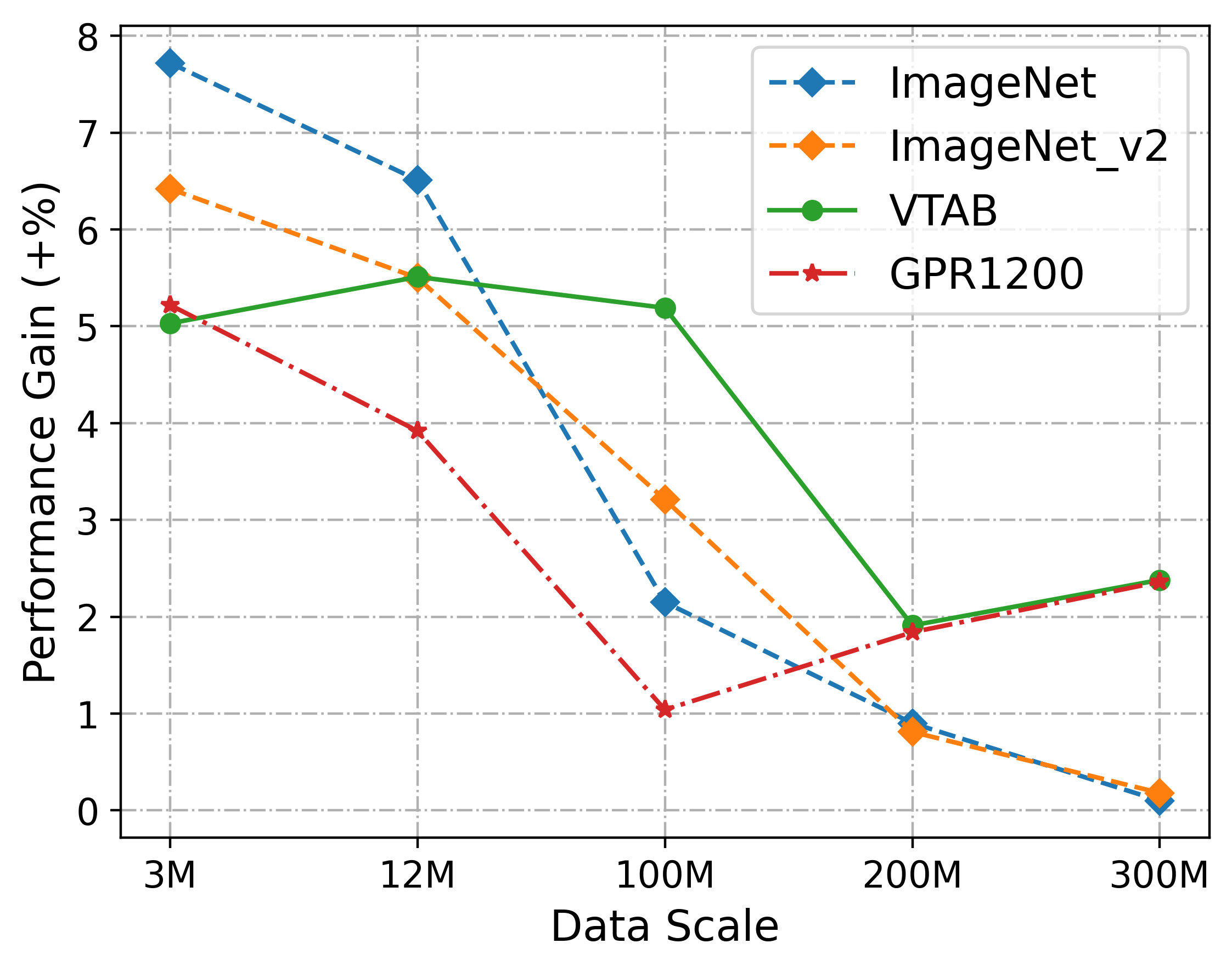}
\vspace{-0.7cm}
\captionof{figure}{\footnotesize{Performance gain on downstream tasks across different data scales.}}
\label{fig:perf_gain}

\end{minipage}

\end{tabular}
\end{table*}

\begin{table*}[t]
    \centering
    \caption{\small{Zero-shot classification accuracy.  Top-1 accuracies (\% ) of VTAB~\citep{vtab} across 9 tasks (6 from natural and 3 from specialized sets) are reported. Full table can be found in Appendix Table A7. 
}}
    \vspace{-0.2cm}
    \resizebox{0.9\textwidth}{!}{
    \begin{tabular}{cc|cccccc|ccc|c}
    \toprule[1.2pt]
    \multirow{2}{*}{\bf Data } & \multirow{2}{*}{\bf Model}  & \multicolumn{6}{c|}{ \bf Natural Sets}  & \multicolumn{3}{c|}{ \bf Specialized Sets } & \multirow{2}{*}{\bf Average } \\
    & &  Caltech101 & CIFAR100 & SVHN & DTD & OxPet & Flowers102 & EuroSAT & RESISC45 & Camelyon \\
    
    \midrule 
    \small 
    \multirow{2}{*}{\bf 3M} & \small CLIP & 39.50 & 9.83 & \bf 20.89 & 7.42 & 7.44 & 10.40 &  \bf 11.94 & 7.93 & 50.65 & 18.45 \\
    &  \small VeCLIP & \bf 54.30 & \bf 17.74 &  18.74 & \bf 11.23 & \bf 10.09 & \bf 22.75 &  7.35 & \bf 16.54 & \bf 52.52 & \bf 23.48 \\
    
    \midrule
    \multirow{2}{*}{\bf 12M} & \small CLIP & 70.43 & 30.06 & \bf 30.11 & 30.69 & 34.51 & 33.67 & 8.87 & 30.05 & 53.46 & 35.76 \\
    & \small VeCLIP & \bf 70.58 & \bf 45.10 &  23.61 & \bf 30.90 & \bf 36.22 & \bf 43.94 & \bf 27.46 & \bf 38.09 & \bf 55.54 & \bf 41.27  \\
    \midrule
    \multirow{2}{*}{\bf 100M} & \small CLIP & 81.44 & 54.75 & 38.70 & 57.28 & \bf 70.51 & 51.71 & 34.45 & 48.56 & 53.87 & 54.59  \\
    & \small VeCLIP & \bf 81.64 & \bf 64.62 & \bf 46.49 & \bf 57.51 &   64.81 & \bf 66.41 & \bf 46.23 & \bf 51.75 & \bf 58.51 & \bf 59.78   \\
    \midrule
    \multirow{2}{*}{\bf 200M} & \small CLIP & 82.30	& 61.87	& 42.83	& \bf 64.29	& \bf 75.60	& 58.67	& 46.73	& \bf 55.59	& 59.30 & 60.79
  \\
    & \small VeCLIP & \bf 83.14 & \bf	68.14 & \bf	44.93	& 61.95	& 72.61	& \bf 68.51	& \bf 47.36	& 55.10	& \bf 62.59	& \bf 62.70
   \\

    \bottomrule[1.2pt]
    \end{tabular}
    }
    \label{tab:zs_classification}
    \vspace{-0.3cm}
\end{table*}

\textbf{Image-to-image retrieval. } We use GPR1200~\cite{gpr1200} with 6 domains for this setting: Google Landmarks V2 (natural and architectural landmarks) denoted as Land, IMDB Faces denoted as Faces, iNat (plants, animals, insects and fungi), INSTRE (planar images and photographs of logos/toys) denoted as INST, ImageNet Sketch denoted as Sketch, and SOP (products and objects, partly isolated). The results (mAP) are summarized in Table~\ref{tab:gpr1200}. We attain a performance gain of 5.22\%/3.92\% under small/medium scales (3M/12M). 
Even upon upscaling the dataset to 200M, we observe a notable 1.84\% increase in average score across six domains. Notably, our primary performance boost is derived from the Sketch domain, underlining the crucial role of visual concepts in zero-shot transferability. Consequently, our visually-enriched captions play a pivotal role in learning such transferability towards downstream tasks.

\textbf{Data efficiency for pre-training.} To show the data efficiency of VeCLIP, we include ALIGN~\cite{ALIGN}, pre-trained on 1.8B data (denoted as 1.8B-ALIGN), and our in-house CLIP~\cite{CLIP} model trained on 1.4B data (denoted as 1.4B-CLIP) as baselines trained at a significantly larger scale. We use these models utilizing over tenfold more data compared to our setting to show the data efficiency of VeCLIP training. 
VeCLIP can outperform 1.4B-CLIP model when scaling up to 100M, representing approximately 7\% of its size, across nearly all downstream tasks. Specifically, in COCO, we achieve +3.44\%/+1.64\% gain in Recall@1 for both retrieval tasks. Upon further scaling to 200M, the improvement becomes even more pronounced, reaching +5.82\%/+3.92\%. 
Furthermore, we achieve a notable +8.60\%/+2.80\% gain in COCO retrieval, as well as a +2.50\%/+0.62\% improvement in Flickr30k, when compared to the 1.8B-ALIGN model. Remarkably, these improvements are achieved with only 11.1\% of the data utilized in the pre-training process.
These results show the data efficacy of VeCLIP. 
When we scale it to 300M, the results are similar to 200M. The results on 300M can be found in Appendix. Therefore, we stop further scaling up the dataset.

\subsection{Results on Image Classification}
\textbf{ImageNet.} We use the same prompt as CLIP (``A photo of a [classname].'') for zero-shot evaluation on both ImageNet~\cite{imagenet} and ImageNetV2~\cite{imagenet_v2}.
The main results are summarized in Table~\ref{tab:imagenet}. We report Top-1, Top-5, and Top-10 accuracies. 
In small and medium-scale settings, we observe a substantial improvement: +10.52\%/+6.42\% gains in Top-1 accuracy on ImageNet/ImageNetV2 under the 3M setting, and +6.51\%/5.50\% gains under the 12M setting. While the improvement becomes marginal upon scaling to 100M/200M, we still achieve noteworthy +2.07\%/+3.21\% and +0.90\%/+0.83\% gains on 100M and 200M across ImageNet and ImageNetV2, respectively. This shows the data efficiency of our pre-training approach.

\textbf{Visual Task Adaptation Benchmark
(VTAB). } Besides ImageNet/ImageNetV2, we also select VTAB~\cite{vtab} for evaluation. Table~\ref{tab:zs_classification} summarizes zero-shot image classification results for both the original CLIP models and our models, utilizing the identical prompt set from CLIP. 
Our approach consistently achieves comparable or superior performance to CLIP across the majority of datasets. For instance, we observe an average accuracy gain of over 5\% under settings of 3M, 12M, and 100M. Even upon scaling up to 200M, we maintain a notable gain of +1.91\%. These results show great robustness on zero-shot classification tasks across different data distributions. We show the overall trend of the performance gain over the data scale in Figure~\ref{fig:perf_gain}.

\subsection{Performance trend across scales }
\begin{figure*}[t!]
\begin{center}
	\begin{tabular}{cc}
	\includegraphics[width=0.48\textwidth]{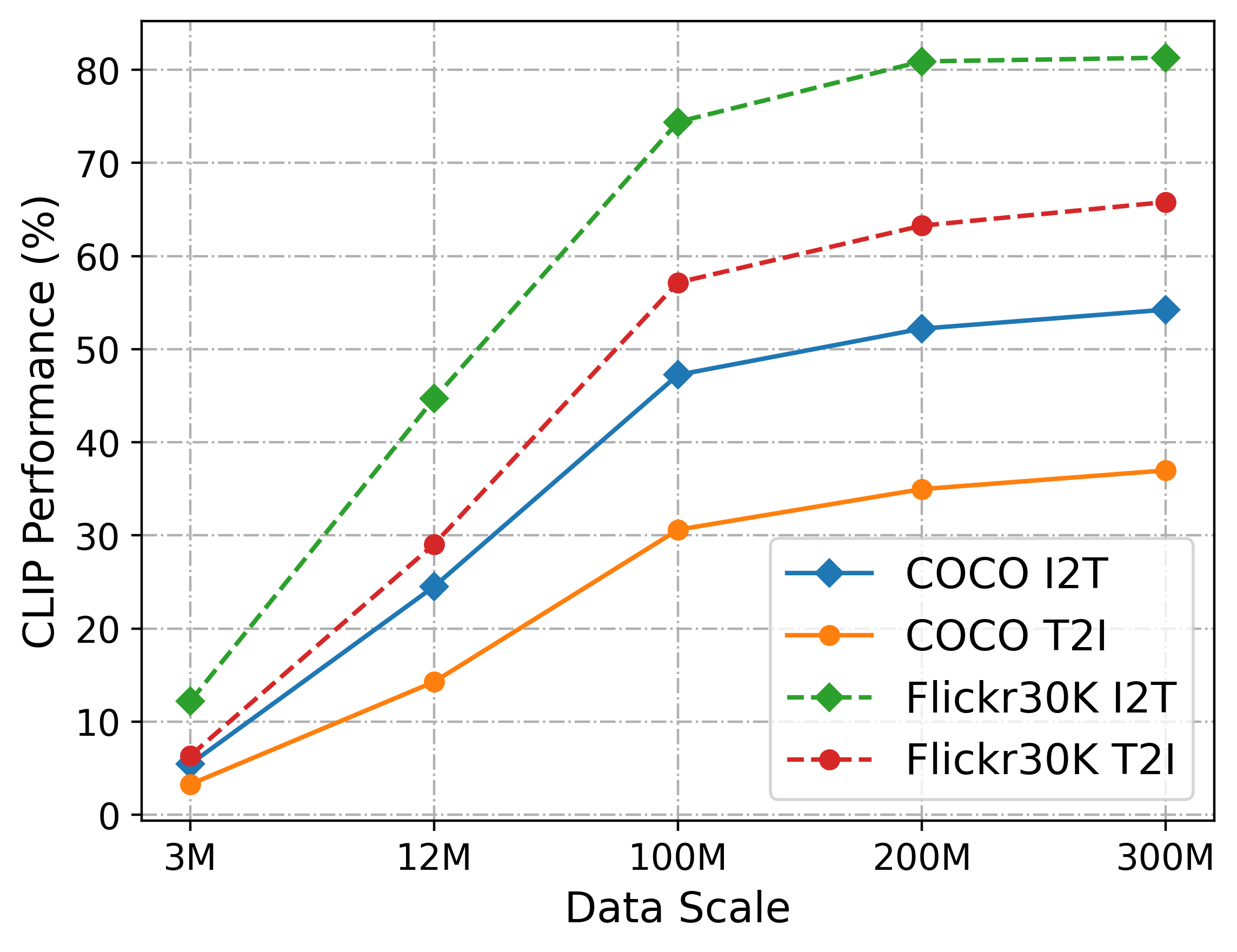} \hspace{-5mm} &
	\includegraphics[width=0.48\textwidth]{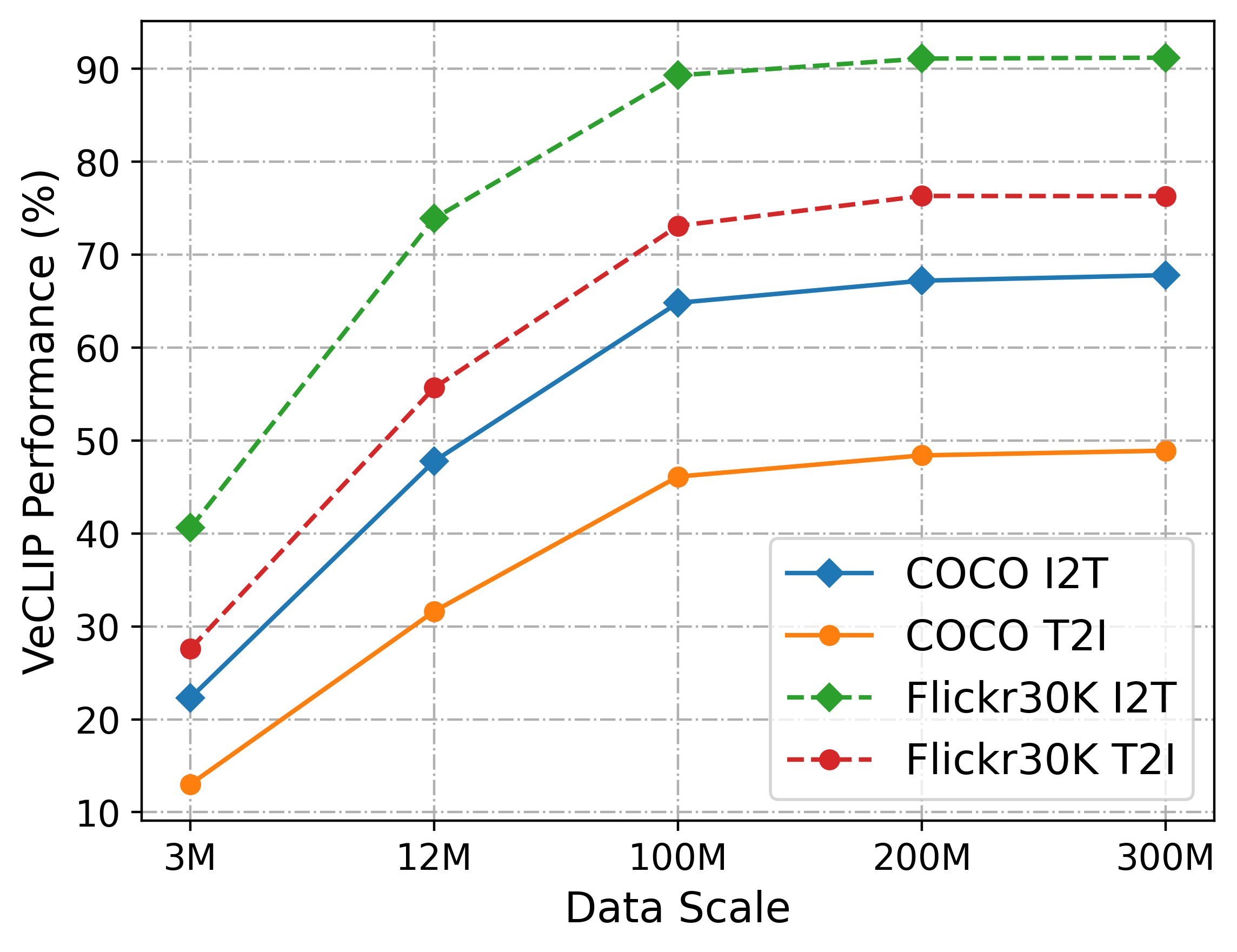} 
 \\
 (a) CLIP & (b) VeCLIP \\
 
	\includegraphics[width=0.48\textwidth]{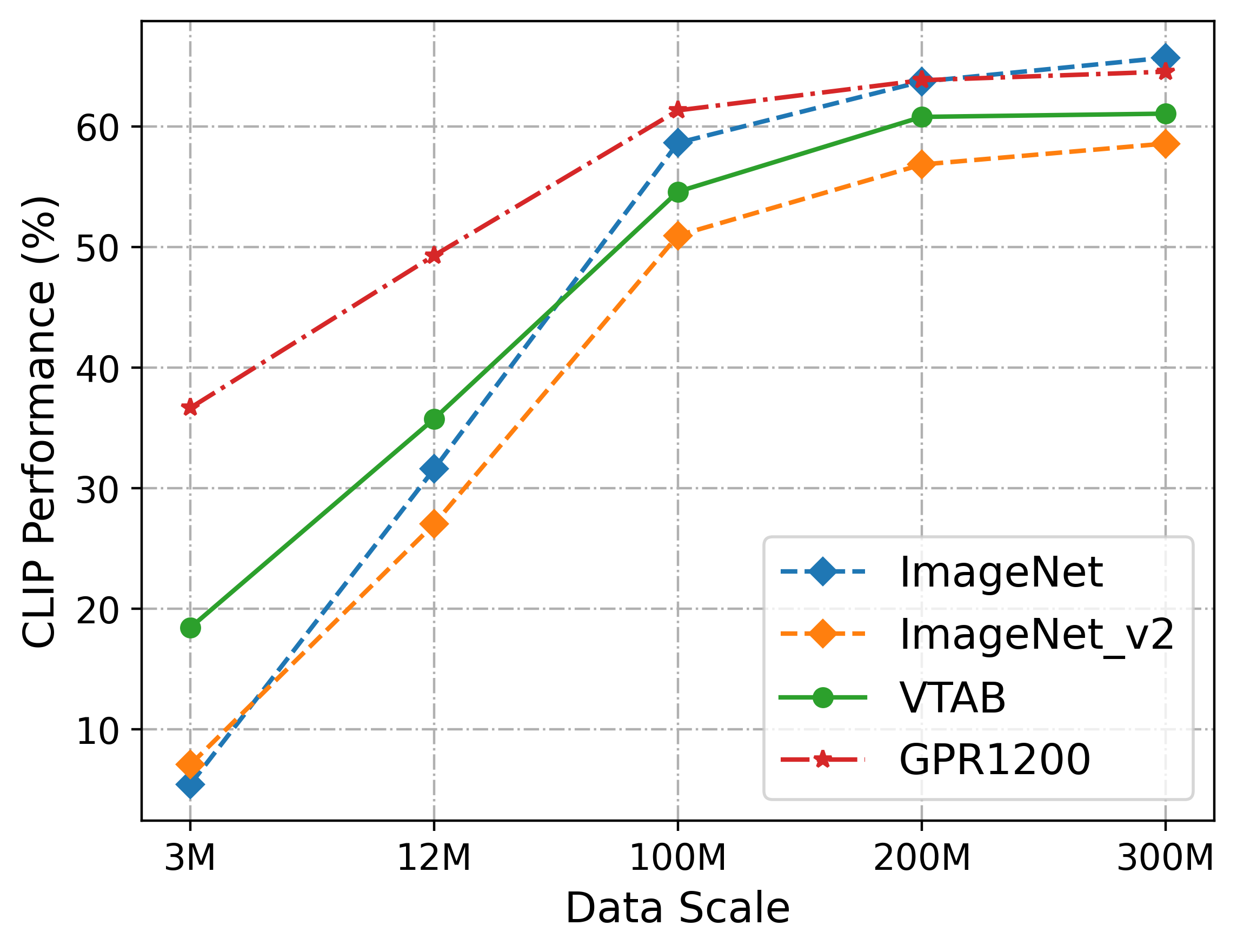} \hspace{-5mm} &
	\includegraphics[width=0.48\textwidth]{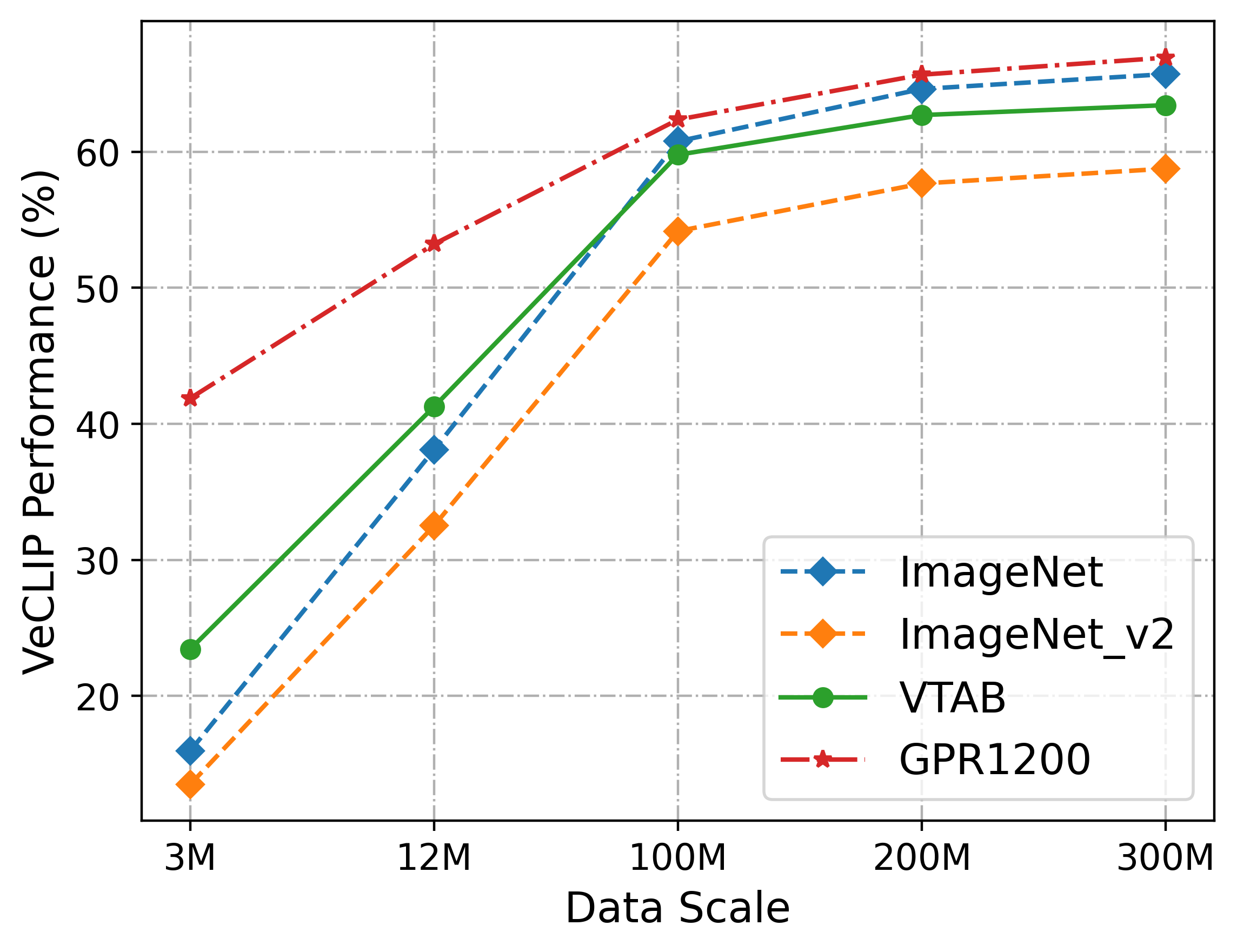}  \\
	    (c) CLIP  & (d) VeCLIP \\
	\end{tabular}
\end{center}
\caption{Performance trend with ViT-B/16 as the vision backbone. (a) and (c) show the trend of CLIP with original AltTexts while (b) and (d) show the trend of VeCLIP with LLM-VeC. The performance is improved significantly when we scale pre-training data up to 100M. Once over 100M, the performance gain becomes gradual and incremental.  }
\label{fig:perf_trend_appd}
\end{figure*}

Besides the performance gain, we also visualize the performance trend across data scales in pre-training. As shown in Figure~\ref{fig:perf_trend_appd}, the performance of CLIP utilizing original AltTexts exhibits a marked surge with the increased data size: while its starting point is poor at 3M, it demonstrates swift progression up to 12M and 100M. However, once scaled beyond 100 million, the performance trend exhibits a gradual and eventually saturated growth. On the other hand, commencing with a higher baseline, VeCLIP employing VeCap demonstrates substantial improvement in comparison to CLIP within small to medium scales (3M and 12M). As we progress beyond 300M, the performance gains of VeCLIP become relatively incremental but still noticeable in retrieval tasks. Both CLIP and VeCLIP reach a saturation point when scaled up to 100M: once over 100M, the performance gain becomes gradual and marginal. 

\subsection{Complementary to other datasets to achieve state-of-the-art performance}
Our VeCap datasets with visual-enriched captions  can also be complementary to other well-curated dataset. For example, DFN~\cite{dfn} has shown benefits on CLIP. To demonstrate that, we train CLIP models with VeCap and DFN separately and also a combination with them. All the models are trained under same configuration for learning rate, maximum steps, and so on. 

We summarize the results in Table~\ref{tab:dfn_veclip}. The high-quality descriptive captions from VeCap can achieve superior results compared to DFN in retrieval tasks. However, the performance on classification tasks are inferior. After we combine DFN and VeCap for training, CLIP can achieve the most improvements for all model sizes.

We also train a H/14 model with resolution 336x336, and compare it with the state-of-the-art models like MetaCLIP~\cite{xu2023demystifying} and DFN~\cite{dfn}. The results are summarized in row 6 to 8 of table \ref{tab:dfn_veclip}. Albeit trained on different resolutions and recipes, the CLIP model with VeCap+DFN is compatible with other models and provide yet another option for downstream tasks~\footnote{Note we took the DFN-H/14 model from its original paper, which is trained 7 epochs, our model is only trained roughly around 2 epochs.}.

\begin{table*}[ht]
    \centering
    \caption{\small{CLIP training with VeCap and DFN~\cite{dfn}, and its comparison with the state-of-the-art models.
}}
    \vspace{-0.2cm}
    \small
    \begin{tabular}{cllccccc}
\toprule[1.2pt]
 \multirow{2}{*}{\bf Model} &  \multirow{2}{*}{\bf Resolution} & \multirow{2}{*}{\bf Data  }  & \multicolumn{2}{c}{ \bf COCO (R@1) } & \multicolumn{2}{c}{\bf Flickr30k (R@1) }  & \multirow{2}{*}{ \bf ImageNet}   \\
\cmidrule(lr){4-5} \cmidrule(lr){6-7}
  &   & &  I2T  & T2I & I2T  & T2I   &        \\
\hline

\multirow{2}{*}{B/16} & \multirow{2}{*}{224} &  DFN~\cite{dfn}    &  63.0 & 43.2  & 87.1 & 70.4   & \textbf{76.2}     \\
& & VeCap+DFN & \textbf{66.3} & \textbf{45.1} & \textbf{88.8} & \textbf{73.6} & \textbf{76.2}  \\

\hline
\rowcolor{tbgray} \multicolumn{8}{c}{\textit{Comparison to other state-of-the-art models}} \\

\rowcolor{tbgray} &  &  DFN~\cite{dfn}   & 68.5 & 48.5 & 89.2 & 75.1 & 81.0 \\
\rowcolor{tbgray} L/14 & 224 & FLIP~\cite{flip}  & 60.2 & 44.2& 89.1& 75.4& 74.6\\
\rowcolor{tbgray}   &  & VeCap+DFN    & \textbf{70.8} & \textbf{49.5}  & \textbf{92.4} & \textbf{78.4} & \textbf{81.1}  \\
 \hline

\rowcolor{tbgray} & 224 &  MetaCLIP~\cite{xu2023demystifying}  &  67.2 & 49.5 & 92.1 & 78.5 & 80.5  \\

\rowcolor{tbgray} H/14 & 378 &  DFN~\cite{dfn}   & 71.8 & 55.6 & 94.0 & 82.1 & {84.4}\\
\rowcolor{tbgray}  & 336 &  VeCap+DFN   & {72.8} & {52.3} & {93.6} & {82.6} & {83.1}  \\

\bottomrule[1.2pt]
\end{tabular}
    \label{tab:dfn_veclip}
\end{table*}
Our VeCLIP with DFN~\cite{dfn} can outperform FLIP~\cite{flip} and OpenAI CLIP with different backbones (as shown in Table A10 in Appendix). Specifically, our ViT-H/14 model achieves impressive 83.1\% of accuracy on ImageNet. We leave the further study of combing the synthetic data~(VeCap) with other data curation approaches as a future work. 


\subsection{Ablation Study}
\definecolor{ForestGreen}{rgb}{0.13, 0.55, 0.13}
\definecolor{Green}{rgb}{0.0, 0.5, 0.0}
\definecolor{green(munsell)}{rgb}{0.0, 0.66, 0.47}
\definecolor{green(ryb)}{rgb}{0.4, 0.69, 0.2}
\definecolor{green(pigment)}{rgb}{0.0, 0.65, 0.31}

\newcommand{\cmark}{\text{\ding{51}}}%
\newcommand{\xmark}{\text{\ding{55}}}%

\newcommand{\tablestyle}[2]{\setlength{\tabcolsep}{#1}\renewcommand{\arraystretch}{#2}\centering}
\newcommand{\km}[1]{{\color{red}[km: #1]}}
\newcommand{\rbg}[1]{{\color{blue}[rbg: #1]}}
\newcommand{\ppd}[1]{{\color{green}[ppd: #1]}}
\newcommand{\bd}[1]{\textbf{#1}}
\newcommand{\app}{\raise.17ex\hbox{$\scriptstyle\sim$}}
\newcommand{\ncdot}{{\mkern 0mu\cdot\mkern 0mu}}

\newcommand{\green}[1]{\multicolumn{1}{c}{\color{green(pigment)}#1}}
\newcommand{\red}[1]{\multicolumn{1}{c}{\color{red}#1}}
\newcommand{\cell}[1]{\multicolumn{1}{r}{#1}}

\begin{table*}[ht]
\centering

\label{table:ablation}

\subfloat[
\footnotesize
Importance of visual-enriched concepts for data quality. We use the AltText with ``Highest CLIP Score'' (HCS) if multiple AltTexts exist on the same image in all settings. 
\label{table:ab_caption}
]{
\centering
\begin{minipage}{0.8\linewidth}{\begin{center}
\resizebox{1.0\textwidth}{!}{
\begin{tabular}{c|ccccccccc}
\toprule[1.2pt]
\multirow{2}{*}{\bf Data} & \multirow{2}{*}{\bf Caption}   & \multirow{2}{*}{\bf \makecell{  Prompt \\ Constraint}  }  & \multicolumn{2}{c}{ \bf \quad COCO (R@1) \quad} & \multicolumn{2}{c}{\bf Flickr30k (R@1)}  & \multirow{2}{*}{ \bf ImageNet}  & \multirow{2}{*}{ \bf ImageNetV2} \\
 &   &   &  I2T  & T2I & I2T  & T2I   &      &   \\

\hline
\multirow{4}{*}{WIT-3M} & AltText  &  -     &  5.18 & 3.40  & 10.50  & 6.88   & 8.02  & \underline{6.88}    \\
& VeC & -  & 16.76 & \bf 9.57 & 32.60 & 20.06 & 7.31 & 6.58 \\
& VeCap  & \red{$\xmark$}  & \underline{17.34} & \underline{9.52} & \underline{37.30} & \underline{21.62} & \underline{8.12} & 6.83 \\
& VeCap  & \green{$\cmark$}  & \bf 18.10 & 9.51 & \bf 40.00 &  \bf 21.94 & \bf 8.20 & \bf 7.39 \\

\hline

\multirow{4}{*}{WIT-12M} & AltText  &  -     &  22.58 & 14.23  & 44.40  & 30.90  & \bf 31.14  & \bf 25.91    \\
& VeC & -  & 40.06 & 24.59 & 64.10 & 43.46 & 7.29 & 14.74 \\
& VeCap  & \red{$\xmark$}  & \underline{44.52} & \bf 27.46 & \underline{70.90} & \underline{50.46} &  \underline{21.05} & 18.11 \\
& VeCap  & \green{$\cmark$}  & \bf 46.82 & \underline{26.61} & \bf 72.60 & \bf 50.94 & 20.99 & \underline{18.41} \\


\bottomrule[1.2pt]
\end{tabular}

}
\end{center}}\end{minipage}
}


\subfloat[
\footnotesize
Importance of the mixed training scheme for data variety. ``HCS'' refers to using the AltText with ``Highest CLIP Score'' while ``random'' refers to randomly selecting one if multiple AltTexts exist.
\label{table:ab_mixed_captio}
]{

\centering
\begin{minipage}{0.95\linewidth}{\begin{center}
\resizebox{0.98\textwidth}{!}{
\begin{tabular}{c|cccccccccc}
\toprule[1.2pt]
\multirow{2}{*}{\bf Data} & \multirow{2}{*}{\bf AltText}   & \multirow{2}{*}{\bf VeCap  } & \multirow{2}{*}{\bf  \makecell{  Training \\ Sampling}  }  & \multicolumn{2}{c}{ \bf \quad COCO (R@1) \quad} & \multicolumn{2}{c}{\bf Flickr30k (R@1)}  & \multirow{2}{*}{ \bf ImageNet}  & \multirow{2}{*}{ \bf ImageNetV2} \\
 &   &  &  &  I2T  & T2I & I2T  & T2I   &      &   \\

\hline
\multirow{5}{*}{WIT-3M} & \green{$\cmark$}   &  \red{$\xmark$} & HCS    &  5.18 & 3.40  & 10.50  & 6.88   & 8.02  & 6.88    \\
& \green{$\cmark$}   &  \red{$\xmark$} & random   & 5.46 & 3.28 & 12.20 & 6.36 & 8.26 & 7.09 \\
& \red{$\xmark$} & \green{$\cmark$} & HCS & 18.10 & 9.51 & \underline{40.00} & 21.94 & 8.20 & 7.39 \\
& \green{$\cmark$}  & \green{$\cmark$} & HCS\&mixed & \underline{19.70} & \underline{12.14} & 39.30 & \underline{25.60} & \underline{14.83} & \underline{12.36} \\
& \green{$\cmark$}  & \green{$\cmark$} & random\&mixed   & \bf 22.30 & \bf 13.01 & \bf 40.60 &  \bf 27.58 & \bf  15.98 & \bf 13.51 \\

\hline
\multirow{5}{*}{WIT-12M} & \green{$\cmark$}   &  \red{$\xmark$} & HCS    &  22.58 & 14.23  & 44.40  & 30.90  &  31.14  &  25.91    \\
& \green{$\cmark$}   &  \red{$\xmark$} & random   & 23.32& 14.28 & 44.70& 29.06 & 31.60 & 27.03 \\
& \red{$\xmark$} & \green{$\cmark$} & HCS & \underline{46.82} & 26.61 & \underline{72.60} & 50.94 & 20.99 & 18.41 \\
& \green{$\cmark$}  & \green{$\cmark$} & HCS\&mixed  & 46.00 & \underline{31.10} & 72.50 & \bf 56.82 & \underline{37.45} & \underline{32.41}\\
& \green{$\cmark$}  & \green{$\cmark$} & random\&mixed    & \bf 47.78 & \bf  31.62 & \bf 73.90 &  \underline{55.68} & \bf 38.11 & \bf 32.51 \\
\hline
\hline
\multirow{2}{*}{CC3M} & \green{$\cmark$}  &  \red{$\xmark$} & -    &  13.88 & 9.64  & 26.30  & 18.04  & 14.59  &  12.52    \\
  & \green{$\cmark$} & \green{$\cmark$} & random\&mixed & \bf 32.04 & \bf 22.07 & \bf 57.20 & \bf 36.54 & \bf 20.73 & \bf 17.90  \\

\bottomrule[1.2pt]
\end{tabular}
}
\end{center}}\end{minipage}
}
\caption{
\small Ablation study of VeCLIP.  The highest score is bold, and the second is underlined. ``mixed'' is our proposed mixed training scheme to alternate among captions. }
\end{table*}

\textbf{Importance of visual-enriched concepts.} 
Different from previous rewriting methods, our primary emphasis lies in fusing visual-enriched concepts extracted from images. 
The ablation findings are summarized in Table~\ref{table:ab_caption}. 
We use 3M/12M as examples to show the performance gain in small/medium scales.
Original AltTexts shows its limitation in retrieval tasks due to its noise and limited image-specific information. VeC generated from LLaVA can boost the performance on retrieval tasks but may hurt the performance on ImageNet zero-shot task. Introducing VeCap can further improve VeCap in all settings. Intriguingly, the zero-shot ImageNet results still lag behind the original AltText. In essence, our VeCap exerts a profound influence on retrieval prowess yet exerts a negative effect on classification tasks. We posit that this phenomenon arises from the following two reasons: 1) there can be a distributional shift in prompts from pre-training to zero-shot inference in ImageNet, particularly noteworthy given the extended length and augmented visual content of VeCap; 2) the data diversity is hurt by LLM rewriting as LLM uses the same writing/paraphrasing style to fuse VeCap and AltText.


\textbf{Importance of mixed training strategies.}  To mitigate the aforementioned issues, we propose a mixed training scheme to alternate between AltTexts and VeCap to provide more data variety during pre-training. We summarize the ablation results of VeCLIP in Table~\ref{table:ab_mixed_captio}. First, we observe a slight performance improvement by randomly selecting one AltText in cases where multiple AltTexts are associated with an image. 
This practice augments data diversity during pre-training. 
Second, interchanging between AltText and VeCap proves to be advantageous, not only in retaining substantial performance gains in retrieval tasks but also in markedly elevating zero-shot results on ImageNet. Lastly, leveraging all AltTexts and VeCap within the mixed training approach in VeCLIP achieves superior results across nearly all settings. 

\textbf{Larger backbone architecture.} We also investigate a larger backbone architecture, e.g., ViT-L/14 and ViT-H/14. The detailed results can be found in both Table~\ref{tab:dfn_veclip} and Appendix C.1.  VeCLIP scaled up in backbone size can consistently outperform the original CLIP in all downstream tasks. Besides, a larger backbone (ViT-L/14) can also achieve up to 5.87\% improvement compared to ViT-B/16. 
These findings support the effectiveness of VeCLIP in improving CLIP pre-training, regardless of the specific underlying backbone architecture.

\textbf{Generalizability of VeCap on well-curated datasets.} Besides our WIT datasets, we evaluate VeCap on well-curated CC3M/CC12M. 
Table~\ref{table:ab_mixed_captio} shows CLIP achieves better performance when pre-trained on CC3M compared to pre-trained on WIT-3M, indicating the importance of high-quality captions for pre-training. With VeCap to further improve the quality of CC3M's captions, CLIP can achieve significant improvement, 
since the captions of CC3M are of higher quality than our noisy WIT dataset.
CC3M with its original captions can outperform the performance of our WIT-3M with AltTexts, indicating CC3M is of higher quality. 
VeCap can significantly improve CLIP under CC3M settings, 
\emph{e.g.}, +18.16\% on the I2T task of COCO and +6.14\% on ImageNet, showing its generalizability on well-curated datasets.  More results  are in Appendix C.2. 

\section{Discussion}
\textbf{Conclusion.} We present a simple yet effective approach to improve CLIP pre-training with leveraging LLaVA and LLMs to rewrite the captions with more visual-enriched concepts. 
VeCLIP is intentionally designed to be scalable and adaptable for handling extensive image-text datasets obtained from web crawling. We conduct a thorough evaluation of VeCLIP on a diverse range of raw and noisy datasets, spanning small, medium, and large scales. The results reveal a substantial performance boost, providing compelling evidence for the effectiveness of our strategy in enhancing large-scale VLM pre-training. VeCLIP can significantly reduce the computational cost and the size of training data for large models to reach competitive results as vanilla CLIP. 

\textbf{Future work.} We employ CLIP as an illustrative instance to highlight the importance of aligning text and images within the training dataset. For future work, we plan to use the collected large-scale dataset to improve the pre-training of other types of VLMs. Further, LLM can generate outputs that encompass factual inaccuracies and hallucinations. Thus, we also plan to delve into more sophisticated filtering techniques to remove such descriptions. 

\textbf{Limitation.} We only leverage LLaVA to exploit the visual concepts. However, the quality measurement metric for such generative AI is still under study. 

{
    \small
    \bibliographystyle{ieeenat_fullname}
    \bibliography{citation}
}

\clearpage
\setcounter{page}{1}
\setcounter{section}{0}

\definecolor{darkred}{RGB}{139, 0, 0}
\definecolor{darkblue}{RGB}{0, 0, 139}
\definecolor{darkgreen}{RGB}{0, 139, 0}
\definecolor{darkyellow}{RGB}{153, 153, 0}
\definecolor{darkpurple}{RGB}{102, 0, 102}
\definecolor{darkbrown}{RGB}{102, 51, 0}

\setcounter{figure}{0}
\makeatletter 
\renewcommand{\thefigure}{A\@arabic\c@figure}
\makeatother
\setcounter{table}{0}
\renewcommand{\thetable}{A\arabic{table}}

\section{Appendix}

We provide additional details for datasets, experimental settings, results, and analysis in the supplementary material. 

\section*{A. Dataset details}
\textbf{Pre-training datasets.}
Instead of using well-curated datasets, we use image-AltText pairs sampled from a web-crawled dataset~\cite{wu2023mofi}. We collect 300M image-text pairs from the Web and denote it as WIT-300M. Based on WIT-300M, we build four subsets to cover from small to large scales. Specifically, 
WIT-200M is a subset of WIT-300M. WIT-100M is a subset of WIT-200M. WIT-12M is a subset of WIT-100M. WIT-3M is a subset of WIT-12M. 



\begin{table*}[ht]
\begin{center}
\caption{Details of 9 VTAB zero-shot classification datasets.}
\vspace{-0.2cm}
\label{table:datasets}
\resizebox{0.7\textwidth}{!}{
\begin{tabular}{@{\hspace{.5em}}l@{\hspace{3.5em}}c@{\hspace{3.5em}}c@{\hspace{2.0em}}c@{\hspace{2.0em}}c@{\hspace{.5em}}}
\toprule[1.2pt]
Dataset        & Metric         & Categories & Train Size & Test Size \\
\midrule
CIFAR-100~\citep{cifar100}      & Accuracy       & 100 & 50,000 & 10,000 \\
SVHN~\citep{SVHN} & Accuracy & 10 & 73,257 & 26,032 \\
DTD~\citep{DTD}            & Accuracy       & 47  & 3,760  & 1,880  \\
Oxford Pets~\citep{oxpet}    & Mean per class & 37  & 3,680  & 3,669  \\
Caltech101~\citep{caltech101}    & Mean per class & 102 & 3,060  & 6,085  \\
Flowers102~\citep{flowers102} & Mean per class & 102 & 2,040  & 6,149  \\
EuroSAT~\citep{EuroSAT}        & Accuracy       & 10  & 10,000 & 5,000  \\
RESISC45~\citep{resisc}       & Accuracy       & 45  & 25,200 & 6,300  \\
Camelyon~\citep{pcam} & Accuracy & 2 & 262,144 & 32,768 \\
\bottomrule[1.2pt]
\end{tabular}
}
\end{center}
\end{table*}
\begin{table*}[ht]
\centering
\caption{
\small Details of the pre-training hyper-parameters for CLIP training on our web-crawled datasets. }
\label{table:hyperparam}
\vspace{-0.2cm}
\subfloat[
\small
Pre-training hyper-parameters on 3M.
\label{table:hyperparam-cc3m}
]{
\centering
\begin{minipage}{0.4\linewidth}{\begin{center}
\resizebox{0.98\textwidth}{!}{
\begin{tabular}{l|l}
\toprule
Config & Value \\
\midrule
Batch size & $8,192$ \\
Optimizer & AdamW \\
Learning rate & $5\times10^{-4}$ \\
Weight decay & $0.5$ \\
Adam $\beta$ & $\beta_1, \beta_2=(0.9, 0.98)$\\
Adam $\epsilon$ & $1\times10^{-8}$ \\
Total epochs & $40$ \\
Warm up epochs & $1$ \\
Learning rate schedule & cosine decay \\
\bottomrule
\end{tabular}}
\end{center}}\end{minipage}
}
\subfloat[
\small
Pre-training hyper-parameters on 12M.
\label{table:hyperparam-cc12m}
]{
\centering
\begin{minipage}{0.4\linewidth}{\begin{center}
\resizebox{0.98\textwidth}{!}{
\begin{tabular}{l|l}
\toprule
Config & Value \\
\midrule
Batch size & $8,192$ \\
Optimizer & AdamW \\
Learning rate & $5\times10^{-4}$ \\
Weight decay & $0.5$ \\
Adam $\beta$ & $\beta_1, \beta_2=(0.9, 0.98)$\\
Adam $\epsilon$ & $1\times10^{-8}$ \\
Total epochs & $35$ \\
Warm up epochs & $1$ \\
Learning rate schedule & cosine decay \\
\bottomrule
\end{tabular}}
\end{center}}\end{minipage}
}
\\
\subfloat[
\small
Pre-training hyper-parameters on 100M.
\label{table:hyperparam-redcaps}
]{
\centering
\begin{minipage}{0.4\linewidth}{\begin{center}
\resizebox{0.98\textwidth}{!}{
\begin{tabular}{l|l}
\toprule
Config & Value \\
\midrule
Batch size & $32,768$ \\
Optimizer & AdamW \\
Learning rate & $5\times10^{-4}$ \\
Weight decay & $0.2$ \\
Adam $\beta$ & $\beta_1, \beta_2=(0.9, 0.98)$\\
Adam $\epsilon$ & $1\times10^{-6}$ \\
Total epochs & $32$ \\
Warm up iterations & $2,000$ \\
Learning rate schedule & cosine decay \\
\bottomrule
\end{tabular}}
\end{center}}\end{minipage}
}
\subfloat[
\small
Pre-training hyper-parameters on 200M.
\label{table:hyperparam-laion}
]{
\centering
\begin{minipage}{0.4\linewidth}{\begin{center}
\resizebox{0.98\textwidth}{!}{
\begin{tabular}{l|l}
\toprule
Config & Value \\
\midrule
Batch size & $32,768$ \\
Optimizer & AdamW \\
Learning rate & $5\times10^{-4}$ \\
Weight decay & $0.2$ \\
Adam $\beta$ & $\beta_1, \beta_2=(0.9, 0.98)$\\
Adam $\epsilon$ & $1\times10^{-6}$ \\
Total epochs & $32$ \\
Warm up iterations & $2,000$ \\
Learning rate schedule & cosine decay \\
\bottomrule
\end{tabular}}
\end{center}}\end{minipage}
}
\\
\end{table*}
\textbf{VTAB datasets. }
We choose 9 classification datasets suitable for zero-shot evaluation from VTAB~\cite{vtab}. Table~\ref{table:datasets} summarizes zero-shot image classification datasets. For both original CLIP models and our models, we use the identical prompt set from CLIP. Every class label is expanded using a collection of prompt templates, as defined by CLIP, including examples like ``A photo of a [classname].'' The class embedding is then computed by taking the average of the embeddings of all such templates, followed by L2-normalization.

\section*{B. Implementation details}
\textbf{Pre-training hyper-parameters.}
We summarize the pre-training hyper-parameters for CLIP training in Table~\ref{table:hyperparam}. We pre-train models on up to 512 TPUs with JAX~\cite{jax2018github}.


\section*{C. More experimental results}
In this section, we present more detailed experimental results and our ablation studies (e.g., generalization of VeCLIP with a large backbone, public and well-curated datasets for pre-training). 

\subsection*{C.1. Larger backbone architectures}
\label{appendix:vit_l}
We also investigate the performance of VeCLIP using a larger backbone architecture, ViT-L/14. The comparison results are summarized in Table~\ref{tab:backbone}. First, VeCLIP shows a consistent improvement over CLIP employing ViT-L/14 across all downstream tasks. Second, VeCLIP utilizing ViT-L/14 surpasses its counterpart employing ViT-B/16, notably excelling in image classification tasks, achieving a notable improvement of over 5\% on both ImageNet and ImageNetV2. This shows that VeCLIP has the potential to be scalable with larger backbone architectures and larger-scale datasets. 
\begin{table*}[ht]
    \centering
    \caption{\small{ Ablation studies on different backbones with VeCLIP. We use 200M as the pre-training dataset.  
}}
    \small
    \resizebox{0.85\textwidth}{!}{
    \begin{tabular}{cccccccc}
\toprule[1.2pt]
 \multirow{2}{*}{\bf Model}   & \multirow{2}{*}{\bf Backbone  }  & \multicolumn{2}{c}{ \bf \quad COCO (R@1) \quad} & \multicolumn{2}{c}{\bf Flickr30k (R@1)}  & \multirow{2}{*}{ \bf ImageNet}  & \multirow{2}{*}{ \bf ImageNetV2} \\
  &   &  I2T  & T2I & I2T  & T2I   &      &   \\

\hline
 CLIP  & ViT-B/16     &  52.20 & 34.97  & 80.90  & 63.23   & 63.72  & 56.84    \\
VeCLIP & ViT-B/16  & \bf 67.20 & \bf 48.40 & \bf 91.10 & \bf 76.32& \bf 64.62 & \bf 57.67 \\
\rowcolor{tbgray}  \multicolumn{2}{c}{\bf Performance Gain}
    & \bf \textcolor{green}{+15.00}  & \bf \textcolor{green}{+13.43} & \bf \textcolor{green}{+10.20} & \bf \textcolor{green}{+13.06}   & \bf \textcolor{green}{+0.90} & \bf \textcolor{green}{+0.81}  \\
\hline
CLIP  & ViT-L/14  & 53.92 & 37.86 & 84.60 & 66.78 & 68.51 & 61.13 \\
VeCLIP & ViT-L/14   & \bf 69.92 & \bf 51.32 & \bf 92.60 & \bf 79.04 & \bf 69.85 & \bf 63.54 \\
\rowcolor{tbgray}  \multicolumn{2}{c}{\bf Performance Gain}
    & \bf \textcolor{green}{+16.00}  & \bf \textcolor{green}{+13.46} & \bf \textcolor{green}{+8.00} & \bf \textcolor{green}{+12.26}   & \bf \textcolor{green}{+1.34} & \bf \textcolor{green}{+2.41}  \\
\hline
\rowcolor{tbgray}  \multicolumn{2}{c}{\bf VeCLIP ViT-L/14 vs B/16}
& \bf \textcolor{green}{+2.72}  & \bf \textcolor{green}{+2.92} & \bf \textcolor{green}{+1.50} & \bf \textcolor{green}{+2.72}   & \bf \textcolor{green}{+5.23} & \bf \textcolor{green}{+5.87}  \\
\bottomrule[1.2pt]
\end{tabular}
    }
    \label{tab:backbone}
\end{table*}

\begin{table*}[ht]
    \centering
    \caption{\small{ Ablation studies on well-curated datasets (CC3M and CC12M~\citep{cc12m}) and the effect of data quality with ViT-B/16 as the vision backbone. 
}}
    \small
    \resizebox{0.85\textwidth}{!}{
    \begin{tabular}{cccccccc}
\toprule[1.2pt]
 \multirow{2}{*}{\bf Model}   & \multirow{2}{*}{\bf Model  }  & \multicolumn{2}{c}{ \bf \quad COCO (R@1) \quad} & \multicolumn{2}{c}{\bf Flickr30k (R@1)}  & \multirow{2}{*}{ \bf ImageNet}  & \multirow{2}{*}{ \bf ImageNetV2} \\
  &   &  I2T  & T2I & I2T  & T2I   &      &   \\

\hline
 \multirow{2}{*}{\bf WIT-3M}  & CLIP    &  5.18 & 3.40  & 10.50  & 6.88   & 8.02  & 6.88   \\
 & VeCLIP  & \bf 22.30 & \bf 13.01 & \bf 40.60 &  \bf 27.58 & \bf  15.98 & \bf 13.51 \\
\rowcolor{tbgray}  \multicolumn{2}{c}{\bf Performance Gain}
    & \bf \textcolor{green}{+17.12}  & \bf \textcolor{green}{+9.61} & \bf \textcolor{green}{+30.10} & \bf \textcolor{green}{+20.70}   & \bf \textcolor{green}{+7.96} & \bf \textcolor{green}{+6.63}  \\
\hline
\multirow{2}{*}{\bf CC3M}   & CLIP  &  13.88 & 9.64  & 26.30  & 18.04  & 14.59  &  12.52 \\
 & VeCLIP   & \bf 32.04 & \bf 22.07 & \bf 57.20 & \bf 36.54 & \bf 20.73 & \bf 17.90 \\
\rowcolor{tbgray}  \multicolumn{2}{c}{\bf Performance Gain}
    & \bf \textcolor{green}{+18.16}  & \bf \textcolor{green}{+12.43} & \bf \textcolor{green}{+30.90} & \bf \textcolor{green}{+18.50}   & \bf \textcolor{green}{+6.14} & \bf \textcolor{green}{+5.38}  \\

\hline
\hline
 \multirow{2}{*}{\bf WIT-12M}  & CLIP    &  22.58 & 14.23  & 44.40  & 30.90  &  31.14  &  25.91   \\
 & VeCLIP  & \bf 47.78 & \bf  31.62 & \bf 73.90 &  \bf 55.68 & \bf 38.11 & \bf 32.51 \\
\rowcolor{tbgray}  \multicolumn{2}{c}{\bf Performance Gain}
    & \bf \textcolor{green}{+25.20}  & \bf \textcolor{green}{+17.39} & \bf \textcolor{green}{+29.50} & \bf \textcolor{green}{+24.78}   & \bf \textcolor{green}{+6.97} & \bf \textcolor{green}{+6.60}  \\
\hline
\multirow{2}{*}{\bf CC12M}   & CLIP  &  37.96 & 24.40  & 59.70  & 44.90  & 39.24  &  34.41 \\
 & VeCLIP   & \bf 53.23 & \bf 36.90 & \bf 75.20 & \bf 62.10 & \bf 45.32 & \bf 40.21 \\
\rowcolor{tbgray}  \multicolumn{2}{c}{\bf Performance Gain}
    & \bf \textcolor{green}{+15.27}  & \bf \textcolor{green}{+12.50} & \bf \textcolor{green}{+15.50} & \bf \textcolor{green}{+17.20}   & \bf \textcolor{green}{+6.08} & \bf \textcolor{green}{+5.80}  \\
    
\bottomrule[1.2pt]
\end{tabular}
    }
    \label{tab:cc3m_appendix}
\end{table*}

\subsection*{C.2. Generalization on well-curated datasets: CC3M and CC12M}
Besides our crawled noisy WIT datasets, we also use a well-curated dataset, e.g., CC3M and CC12M~\cite{cc12m}, to show the effectiveness and generalizability of our proposed approach on well-curated datasets. CC3M and CC12M~\cite{cc12m} were curated via several rounds of comprehensive refining and filtering to get high-quality image-caption pairs. We show high-quality examples of CC3M  and the comparison of CC3M's captions and WIT-3M's AltTexts in Appendix D. 
We present an experimental comparison between our crawled WIT datasets and well-curated CC3M/CC12M~\cite{cc12m} in this subsection. 

\textbf{3M.} As shown in Table~\ref{tab:cc3m_appendix}, CC3M outperforms WIT-3M when coupled with CLIP pre-training, yielding a notable increase of +10.70\% on the COCO I2T task. Additionally, VeCLIP exhibits substantial improvement for both WIT-3M and CC3M. Notably, we achieve a remarkable over 30\% improvement on the I2T task in Flickr30K, and an impressive over 5\% boost on ImageNet and ImageNetV2. 

\textbf{12M.} Similar to 3M settings, CC12M exhibits superior quality and attains better results in contrast to WIT-12M when utilized with CLIP and original AltTexts. VeCLIP demonstrates notable improvements for both WIT-12M and CC12M. For instance, VeCLIP yields a remarkable +12.27\% increase in the I2T task of COCO, along with an impressive over 5\% improvement on both ImageNet and ImageNetV2. These findings emphasize the effectiveness and generalizability of VeCLIP in both noisy web-crawled datasets and meticulously curated datasets, where a richer set of visual concepts is harnessed for pre-training.



\subsection*{C.3. Complete visual descriptions vs simplified entity representations}
In Table~\ref{table:ab_mixed_captio} of the main paper, we note that sole training on VeCap might detriment zero-shot performance in comparison to the original AltText. Conversely, our mixed training approach yields optimal outcomes. This intriguing finding propels us toward a more profound investigation of zero-shot classification tasks. Following established works~\cite{CLIP,fan2023improveclip}, we employ an identical set of prompting templates, such as ``a photo of a [CLS]'' for ImageNet~\cite{imagenet}. It is conceivable that this direct and uncomplicated prompt may diverge significantly from VeCap's pre-training, which encompasses a more extensive and intricate set of visual concepts. To address this, we reformulate VeCap into a format as  Simplified Entity Representation (SER). Specifically, we employ the NLTK package to extract entities from VeCap and subsequently apply filtering to retain only noun entities, denoted as $(A, B, C...) \in U$. This transformation results in VeCap being presented as ``a photo of [$U$]'', offering a concise representation of all extracted entities. The results are summarized in Table~\ref{tab:ser}. 
Surprisingly, we find that even with SER-style captions, the zero-shot performance remains inferior to that achieved with the original AltText. We hypothesize that this discrepancy may arise from a lack of data diversity. When all sentences adhere to the same distribution, there exists a risk of overfitting in the pre-trained model, resulting in suboptimal performance in downstream tasks.

\begin{table*}[ht]
    \centering
    \caption{\small{ Ablation studies on VeCap and Simplied Entities Representation (SER). We use ViT-B/16 as the backbone and use 200M as the pre-trained dataset.  
}}
    \small
    \resizebox{0.85\textwidth}{!}{
    \begin{tabular}{cccccccc}
\toprule[1.2pt]
 \multirow{2}{*}{\bf Model}   & \multirow{2}{*}{\bf Caption  }  & \multicolumn{2}{c}{ \bf \quad COCO (R@1) \quad} & \multicolumn{2}{c}{\bf Flickr30k (R@1)}  & \multirow{2}{*}{ \bf ImageNet}  & \multirow{2}{*}{ \bf ImageNetV2} \\
  &   &  I2T  & T2I & I2T  & T2I   &      &   \\

\hline
 CLIP  & AltText     &  52.20 & 34.97  & 80.90  & 63.23   & 63.72  & 56.84    \\
 VeCLIP & SER & 65.88 & \bf 49.04 & 89.20 & 75.96 & 58.58 & 52.89 \\  
\rowcolor{tbgray} VeCLIP & VeCap  & \bf 67.20 & 48.40 & \bf 91.10 & \bf 76.32& \bf 64.62 & \bf 57.67 \\

\bottomrule[1.2pt]
\end{tabular}
    }
    \label{tab:ser}
\end{table*}

\subsection*{C.4. Main results with WIT-300M }
We show the detailed results with the Web-crawled Image-Text 300M dataset (WIT-300M) here. We summarize the results on various downstream tasks in Table~\ref{tab:imagenet_appendix}. There are two major observations. First, we observe that the results obtained with a dataset size of 300M are close to those achieved with 200M for both CLIP and VeCLIP models. This suggests that a dataset scale of 200 million is sufficient for effectively training a ViT-B/16-based CLIP model. Second, VeCLIP achieves significant improvement on retrieval tasks even under 300M settings. Nevertheless, the improvement observed in ImageNet/ImageNetV2 is marginal.

\begin{table*}[ht]
    \centering
    \caption{\small{Zero-shot classification accuracy.  Top-1 Accuracies (\% ) of VTAB~\citep{vtab} across 9 tasks (6 from natural and 3 from specialized sets) are reported. 
}}
    \vspace{-0.2cm}
    \resizebox{1.0\textwidth}{!}{
    \begin{tabular}{cc|cccccc|ccc|c}
    \toprule[1.2pt]
    \multirow{2}{*}{\bf Data } & \multirow{2}{*}{\bf Model}  & \multicolumn{6}{c|}{ \bf Natural Sets}  & \multicolumn{3}{c|}{ \bf Specialized Sets } & \multirow{2}{*}{\bf Average } \\
    & &  Caltech101 & CIFAR100 & SVHN & DTD & OxPet & Flowers102 & EuroSAT & RESISC45 & Camelyon \\
    
    \midrule 
    \multicolumn{12}{c}{\bf \textit{Model Architecture: ViT-B/16}}\\
    \midrule 
    \small 
    \multirow{2}{*}{\bf 3M} & \small CLIP & 39.50 & 9.83 & \bf 20.89 & 7.42 & 7.44 & 10.40 &  \bf 11.94 & 7.93 & 50.65 & 18.45 \\
    &  \small VeCLIP & \bf 54.30 & \bf 17.74 &  18.74 & \bf 11.23 & \bf 10.09 & \bf 22.75 &  7.35 & \bf 16.54 & \bf 52.52 & \bf 23.48 \\
    \rowcolor{tbgray}  \multicolumn{2}{c|}{\bf Performance Gain}
    & \bf \textcolor{green}{+14.80}  & \bf \textcolor{green}{+7.91}
    & \bf \textcolor{red}{-2.15}  & \bf \textcolor{green}{+3.81}
    & \bf \textcolor{green}{+2.65}  & \bf \textcolor{green}{+12.35}
    & \bf \textcolor{red}{-4.59}  & \bf \textcolor{green}{+8.61}
    & \bf \textcolor{green}{+1.87}  & \bf \textcolor{green}{+5.03} \\
 
    \midrule

    \multirow{2}{*}{\bf 12M} & \small CLIP & 70.43 & 30.06 & \bf 30.11 & 30.69 & 34.51 & 33.67 & 8.87 & 30.05 & 53.46 & 35.76 \\
    & \small VeCLIP & \bf 70.58 & \bf 45.10 &  23.61 & \bf 30.90 & \bf 36.22 & \bf 43.94 & \bf 27.46 & \bf 38.09 & \bf 55.54 & \bf 41.27  \\
    \rowcolor{tbgray}  \multicolumn{2}{c|}{\bf Performance Gain}
    & \bf \textcolor{green}{+0.15}  & \bf \textcolor{green}{+15.04}
    & \bf \textcolor{red}{-6.50}  & \bf \textcolor{green}{+0.21}
    & \bf \textcolor{green}{+1.71}  & \bf \textcolor{green}{+10.27}
    & \bf \textcolor{green}{+18.59}  & \bf \textcolor{green}{+8.04}
    & \bf \textcolor{green}{+2.08}  & \bf \textcolor{green}{+5.51} \\
    \midrule
    \multirow{2}{*}{\bf 100M} & \small CLIP & 81.44 & 54.75 & 38.70 & 57.28 & \bf 70.51 & 51.71 & 34.45 & 48.56 & 53.87 & 54.59  \\
    & \small VeCLIP & \bf 81.64 & \bf 64.62 & \bf 46.49 & \bf 57.51 &   64.81 & \bf 66.41 & \bf 46.23 & \bf 51.75 & \bf 58.51 & \bf 59.78   \\
    \rowcolor{tbgray}  \multicolumn{2}{c|}{\bf Performance Gain}
    & \bf \textcolor{green}{+0.20}  & \bf \textcolor{green}{+9.87}
    & \bf \textcolor{green}{+7.79}  & \bf \textcolor{green}{+0.23}
    & \bf \textcolor{red}{-5.70}  & \bf \textcolor{green}{+14.70}
    & \bf \textcolor{green}{+11.78}  & \bf \textcolor{green}{+3.19}
    & \bf \textcolor{green}{+4.64}  & \bf \textcolor{green}{+5.19} \\
    \midrule
    \multirow{2}{*}{\bf 200M} & \small CLIP & 82.30	& 61.87	& 42.83	& \bf 64.29	& \bf 75.60	& 58.67	& 46.73	& \bf 55.59	& 59.30 & 60.79
  \\
    & \small VeCLIP & \bf 83.14 & \bf	68.14 & \bf	44.93	& 61.95	& 72.61	& \bf 68.51	& \bf 47.36	& 55.10	& \bf 62.59	& \bf 62.70
   \\
   \rowcolor{tbgray}  \multicolumn{2}{c|}{\bf Performance Gain}
    & \bf \textcolor{green}{+0.84}  & \bf \textcolor{green}{+6.27}
    & \bf \textcolor{green}{+2.10}  & \bf \textcolor{red}{-2.34}
    & \bf \textcolor{red}{-2.99}  & \bf \textcolor{green}{+9.84}
    & \bf \textcolor{green}{+0.63}  & \bf \textcolor{red}{-0.49}
    & \bf \textcolor{green}{+3.29}  & \bf \textcolor{green}{+1.91} \\

    \midrule
    \multirow{2}{*}{\bf 300M} & \small CLIP & \bf 83.58	& 63.36	& 50.04	& \bf 66.16	&  74.30	& 61.81	& 39.95	& 56.44	& \bf 53.94 & 61.06
  \\
    & \small VeCLIP &  83.07 & \bf	68.37 & \bf	50.07	& 65.98	& \bf 75.36	& \bf 69.71	& \bf 48.28	& \bf 58.09	&  51.94	& \bf 63.43
   \\
   \rowcolor{tbgray}  \multicolumn{2}{c|}{\bf Performance Gain}
    & \bf \textcolor{red}{-0.51}  & \bf \textcolor{green}{+5.01}
    & \bf \textcolor{green}{+0.03}  & \bf \textcolor{red}{-0.18}
    & \bf \textcolor{green}{1.06}  & \bf \textcolor{green}{+7.90}
    & \bf \textcolor{green}{+8.33}  & \bf \textcolor{green}{+1.65}
    & \bf \textcolor{red}{-2.00}  & \bf \textcolor{green}{+2.37} \\

    \bottomrule[1.2pt]
    \end{tabular}
    }
    \label{tab:vtab_appendix}
\end{table*}
\begin{table*}[ht]
\begin{tabular}{cc}

\begin{minipage}[t]{0.49\linewidth}
\vspace{0pt}
\centering

\caption{\small{Image-to-Image retrieval results (mAP) on 6-domain GPR1200. 
}}
\vspace{-0.2cm}
\label{tab:gpr1200_appendix}   
\small
\resizebox{1.0\textwidth}{!}{
\begin{tabular}{cc|cccccc|g}
\toprule[1.2pt]
\multirow{2}{*}{\bf Data} & \multirow{2}{*}{\bf Model} & \multicolumn{6}{c|}{\bf Domain Name}  & \\

& & Land & Faces & iNat   & INST & Sketch & SOP & \bf  All \\
\midrule 
\small 
\multirow{2}{*}{\bf 3M} & \small CLIP & 57.98 & 20.76& 17.61 & 31.14	& 18.23	& 74.29 & 36.67 \\
&  \small VeCLIP  & \bf 66.55	& \bf 23.51	&  \bf20.43 & \bf 38.63	& \bf 24.59	& \bf 77.65 & \bf 41.89\\

\midrule
\multirow{2}{*}{\bf 12M} & \small CLIP & 74.47 & 30.65 & 23.60 & 52.15	& 30.68	& 84.25 & 49.30 \\
&  \small VeCLIP  & \bf 79.30	& \bf 31.72 & \bf 25.53 & \bf 56.65	& \bf 41.42	& \bf 84.69 & \bf 53.22\\

\midrule
\multirow{2}{*}{\bf 100M} & \small CLIP & \bf 85.64 & \bf 51.68 & 29.66 & 68.19	& 42.45	& 90.38 & 61.33 \\
&  \small VeCLIP  & 85.59 	&  42.83 & \bf 30.72 & \bf 71.96	& \bf 52.59	& \bf 90.54 & \bf 62.37\\

\midrule
\multirow{2}{*}{\bf 200M} & \small CLIP & \bf 86.96 &\bf 56.54 & 30.95 & 71.51	& 46.03	& 90.95 & 63.83 \\
&  \small VeCLIP  & 86.40 	&  48.48 & \bf 31.72 & \bf 73.74	& \bf 56.52	& \bf 91.16 & \bf 65.67 \\

\midrule
\multirow{2}{*}{\bf 300M} & \small CLIP & \bf 87.17 &\bf 57.09 & 31.83& 72.80	& 47.03	& \bf 91.30 & 64.54 \\
&  \small VeCLIP  & 86.22 	&  48.51 & \bf 32.05 & \bf 75.29	& \bf 56.18	&  91.25 & \bf 66.91 \\

\bottomrule[1.2pt]
\end{tabular}
}

\end{minipage}
 & 
\begin{minipage}[t]{0.45\linewidth}
\vspace{0pt}
\centering
\caption{\small{Zero-shot classification results (Top-$k$ Accuracy) on ImageNet and ImageNetV2.
}}
\vspace{-0.2cm}
\label{tab:imagenet_appendix}   
\small
\resizebox{1.0\textwidth}{!}{
\begin{tabular}{cc|ccc|ccc}
\toprule[1.2pt]
\multirow{2}{*}{\bf Data } & \multirow{2}{*}{\bf Model} & \multicolumn{3}{c|}{ \bf ImageNet} & \multicolumn{3}{c}{ \bf ImageNetV2} \\
& & Top-1 & Top-5 & Top-10   & Top-1 & Top-5 & Top-10  \\
\midrule 
\small 
\multirow{2}{*}{\bf 3M} & \small CLIP & 5.46 & 21.05 & 28.70 & 7.09	& 18.52	& 25.83 \\
&  \small VeCLIP  & \bf 15.98	& \bf 34.11	& \bf 43.23 & \bf 13.51	& \bf 30.03	& \bf 38.93 \\

\midrule
\multirow{2}{*}{\bf 12M} & \small CLIP & 31.60	& 58.80	& 69.49 & 27.03 &	52.68	& 63.37 \\
&  \small VeCLIP  & \bf 38.11	& \bf 66.74	& \bf 76.36 & \bf 32.53	& \bf 60.16 &	\bf 70.50 \\

\midrule
\multirow{2}{*}{\bf 100M} & \small CLIP & 58.64	& 85.82 &	91.79 & 50.96	& 79.77	& 86.91 \\
&  \small VeCLIP  & \bf 60.77	& \bf 87.77	& \bf 93.16 & \bf 54.17	& \bf 82.51 &	\bf 89.24 \\

\midrule
\multirow{2}{*}{\bf 200M} & \small CLIP & 63.72	& 89.26 & 94.11 & 56.84 & 83.50 & 89.79 \\
&  \small VeCLIP  & \bf 64.62	& \bf 90.27	& \bf 94.90 & \bf 57.67	& \bf 85.24 &	\bf 91.62 \\

\midrule
\multirow{2}{*}{\bf 300M} & \small CLIP & 65.70	& 90.55 & 94.87 & 58.58 & 85.32 & 91.35 \\
&  \small VeCLIP  & \bf 65.71	& \bf 91.15	& \bf 95.36 & \bf 58.76& \bf 86.31 &	\bf 91.95 \\

\bottomrule[1.2pt]
\end{tabular}
}

\end{minipage}

\end{tabular}
\end{table*}


As shown in Table~\ref{tab:sota}, our VeCLIP with DFN~\cite{dfn} can outperform FLIP~\cite{flip} and OpenAI CLIP with different backbones. Specifically, our ViT-H/14 model achieves impressive 83.1\% of accuracy on ImageNet. We leave the further study of combing the synthetic data~(VeCap) with other data curation approaches as a future work. 
\begin{table*}[ht]
    \centering
    \caption{\small{Comparison bwetween VeCLIP and other models.   
}}
    \vspace{-0.2cm}
    \small
    \resizebox{0.85\textwidth}{!}{
    \begin{tabular}{cccccccc}
\toprule[1.2pt]
\multirow{2}{*}{\bf Backbone} &  \multirow{2}{*}{\bf Model}   & \multirow{2}{*}{\bf Data  }  & \multicolumn{2}{c}{ \bf \quad COCO (R@1) \quad} & \multicolumn{2}{c}{\bf Flickr30k (R@1)}  & \multirow{2}{*}{ \bf ImageNet}   \\
&  &   &  I2T  & T2I & I2T  & T2I   &         \\

\hline
\multirow{3}{*}{ViT-B/16}   & OpenAI CLIP & OpenAI-400M & 53.8 & 33.1 & 88.0 & 68.7 & 68.6 \\
& FLIP~\cite{flip} & LAION-400M & - & - & - & - & 68.0 \\
  & VeCLIP & DFN~\cite{dfn} + VeCap & \textbf{66.3} & \textbf{45.1} & \textbf{88.8} & \textbf{73.6} & \textbf{76.2} \\
\hline
\multirow{3}{*}{ViT-L/14} &  OpenAI CLIP & OpenAI-400M & 58.4 & 37.8 & 88.0 & 68.7 & 75.3 \\ 
& FLIP~\cite{flip} & LAION-400M & 60.2 & 44.2& 89.1& 75.4& 74.6\\
& VeCLIP & DFN~\cite{dfn} + VeCap & \textbf{71.1} & \textbf{51.1} & \textbf{93.1} & \textbf{81.0} & \textbf{82.0} \\
\hline
\multirow{1}{*}{ViT-H/14} 
& VeCLIP & DFN~\cite{dfn} + VeCap & \textbf{72.8} & \textbf{52.3} & \textbf{93.6} & \textbf{82.6} & \textbf{83.1} \\

\bottomrule[1.2pt]
\end{tabular}
    }
    \label{tab:sota}
\end{table*}

\section*{D. Caption quality comparison between well-curated Datasets and WIT datasets}
In Appendix C.2, we find CLIP performs notably better when pre-trained on CC3M compared to the case of being pre-trained on noisy crawled WIT datasets due to several rounds of filtering and refining involved in the curation of CC3M and CC12M. In this section, we show detailed captions from CC3M and compare them with AltTexts from WIT datasets. \\

\noindent{\bf Here we provide more examples of AltText and LLM-VeC from WIT-3M:}
\begin{enumerate}[noitemsep, leftmargin=*]

\item \textbf{AltText:} \textcolor{darkblue}{Ring Capri Pomellato | Pomellato Online Boutique} \\
\textbf{VeCap:} \textcolor{darkgreen}{Pomellato's Ring Capri features a delicate and elegant white stone or possibly three pearls, set against a white background.}

\item \textbf{AltText:} \textcolor{darkblue}{Fiamma F45 L 450 Royal Blue Awning.} \\
\textbf{VeCap:} \textcolor{darkgreen}{The Fiamma F45 L 450 Royal Blue Awning is featured on a white car with a visible red logo for perfect closing, parked in a driveway under a tree, with a house in the background.} 

\item \textbf{AltText:} \textcolor{darkblue}{Union votes for strike on pensions} \\
\textbf{VeCap:} \textcolor{darkgreen}{The man with white hair, dressed in a suit and tie, exhibits a surprised or expressive look on his face, with his mouth open and hand near his face, creating a dynamic and energetic expression.} 

\item \textbf{AltText:} \textcolor{darkblue}{r/reallifedoodles - I can show you the world} \\
\textbf{VeCap:} \textcolor{darkgreen}{The large orange and black drone hovers in the air, carrying two small teddy bears attached to it, above a patio area, as seen in the image.} 

\item \textbf{AltText:} \textcolor{darkblue}{20 Amazon Skincare Products That Keep Selling Out} \\
\textbf{VeCap:} \textcolor{darkgreen}{20 Amazon skincare products that keep selling out feature a happy woman with dark skin, wearing a white shirt and covering her face with her hands, with a white spot or patch on her skin.} 


\item \textbf{AltText:} \textcolor{darkblue}{Durable White Arcane Dining Console Table With 6 Hidden Chairs} \\
\textbf{VeCap:} \textcolor{darkgreen}{A durable white arcane dining console table with 6 hidden chairs is visually appealing and ready for use, as seen in the image featuring a dining set with a white table and two benches, surrounded by black chairs.
} 


\item \textbf{AltText:} \textcolor{darkblue}{Peaceful apartment with wi fi internet access, near old Quebec.} \\
\textbf{VeCap:} \textcolor{darkgreen}{Experience a peaceful stay in a cozy apartment with Wi-Fi internet access, located near historic Old Quebec, featuring a charming dining room with a set table and chairs on a hardwood floor, complete with a white refrigerator in the background.
} 

\item \textbf{AltText:} \textcolor{darkblue}{CABLE BUJIA CHEVROLET CORSA 1.0 1.4 EFI FERRAZZI CABLE BUJIA CHEVROLET CORSA 1.0 1.4 EFI FERRAZZI} \\
\textbf{VeCap\:} \textcolor{darkgreen}{An array of cords and wires, comprising a black rubber cable, is displayed on a pristine surface, featuring diverse configurations and orientations, with some lying horizontally and others positioned at angles.
}

\end{enumerate}

\noindent{\bf Here we provide more examples of original caption and VeCap from CC3M:}
\vspace{-2mm}
\begin{enumerate}[noitemsep, leftmargin=*]

\item \textbf{CC3M Caption:} \textcolor{darkblue}{person runs with the ball during their training session on friday.} \\
\textbf{VeCap:} \textcolor{darkgreen}{A group of soccer players, clad in red and black jerseys, are energetically engaging in a game on a vast field, with some running and others immersed in the action, dispersed across the terrain.}

\item \textbf{CC3M Caption:} \textcolor{darkblue}{a house with red roof with some bushes and a lamp post in front.} \\
\textbf{VeCap:} \textcolor{darkgreen}{A prominent two-story beige building with a distinctive tile roof stands out in the area, illuminated by a nearby lamp post. The building appears to be a complex with several houses or apartments, adding a touch of complexity to the surroundings.} 

\item \textbf{CC3M Caption:} \textcolor{darkblue}{eating a big sweet cupcake with chocolate at cafe.} \\
\textbf{VeCap:} \textcolor{darkgreen}{A person holds a half-eaten blueberry muffin on a plate, standing next to a dining table with a cup, while eating a big sweet cupcake with chocolate at a cafe.} 

\item \textbf{CC3M Caption:} \textcolor{darkblue}{paper heart with red ribbon and a bow.} \\
\textbf{VeCap:} \textcolor{darkgreen}{A pink background showcases a heart-shaped box with a bow, adorned in white with the message ``Happy Valentine's Day,'' positioned centrally within the image.} 

\item \textbf{CC3M Caption:} \textcolor{darkblue}{person andactor at the premiere} \\
\textbf{VeCap:} \textcolor{darkgreen}{Two individuals, a man and a woman, are depicted standing together, both attired in formal attire. The man is donning a tuxedo with a black bow tie, while the woman is wearing a long dress. They seem to be positioning themselves for a photograph, possibly at a formal event.}

\item \textbf{CC3M Caption:} \textcolor{darkblue}{wedding ceremony on the beach} \\
\textbf{VeCap:} \textcolor{darkgreen}{A picturesque wedding ceremony unfolds on a stunning white sandy beach, where perfectly arranged chairs accommodate guests in formal attire. The groom and bride exude joy and love, basking in the warm sunlight.
}

\item \textbf{CC3M Caption:} \textcolor{darkblue}{revenge is a dish best served cold ... with lots of lettuce .} \\
\textbf{VeCap:} \textcolor{darkgreen}{A large, possibly turtle, tortoise with an angry expression sits on rocks, displaying a saying or text message that reads ``Revenge is a dish best cold served with lots of lettuce.''
} 

\item \textbf{CC3M Caption:} \textcolor{darkblue}{interior of an abandoned factory} \\
\textbf{VeCap:} \textcolor{darkgreen}{The sunlit interior of an industrial building stands in contrast to its darker exterior, with numerous windows allowing natural light to flood the space, giving it an empty and open appearance devoid of people or personal touches.
}

\end{enumerate}

Examining the aforementioned instances, it becomes evident that CC3M's captions exhibit a notable level of precision and high quality, displaying a closer alignment with the corresponding images. Conversely, WIT-3M's AltTexts tend to be more cluttered, signaling a comparatively subpar performance in contrast to CC3M. Upon implementing VeCap, even though CC3M's captions are of high quality, they are enhanced with more visual concepts leveraged via VeCap. 
Such integration of enriched visual concepts accounts for the significant improvement we achieve in retrieval tasks (the results are shown in Table~\ref{tab:cc3m_appendix}).



\section*{E. More examples of WIT with VeCap}
\begin{figure*}[ht]
\begin{center}
	\begin{tabular}{c}
	\includegraphics[width=0.87\textwidth]{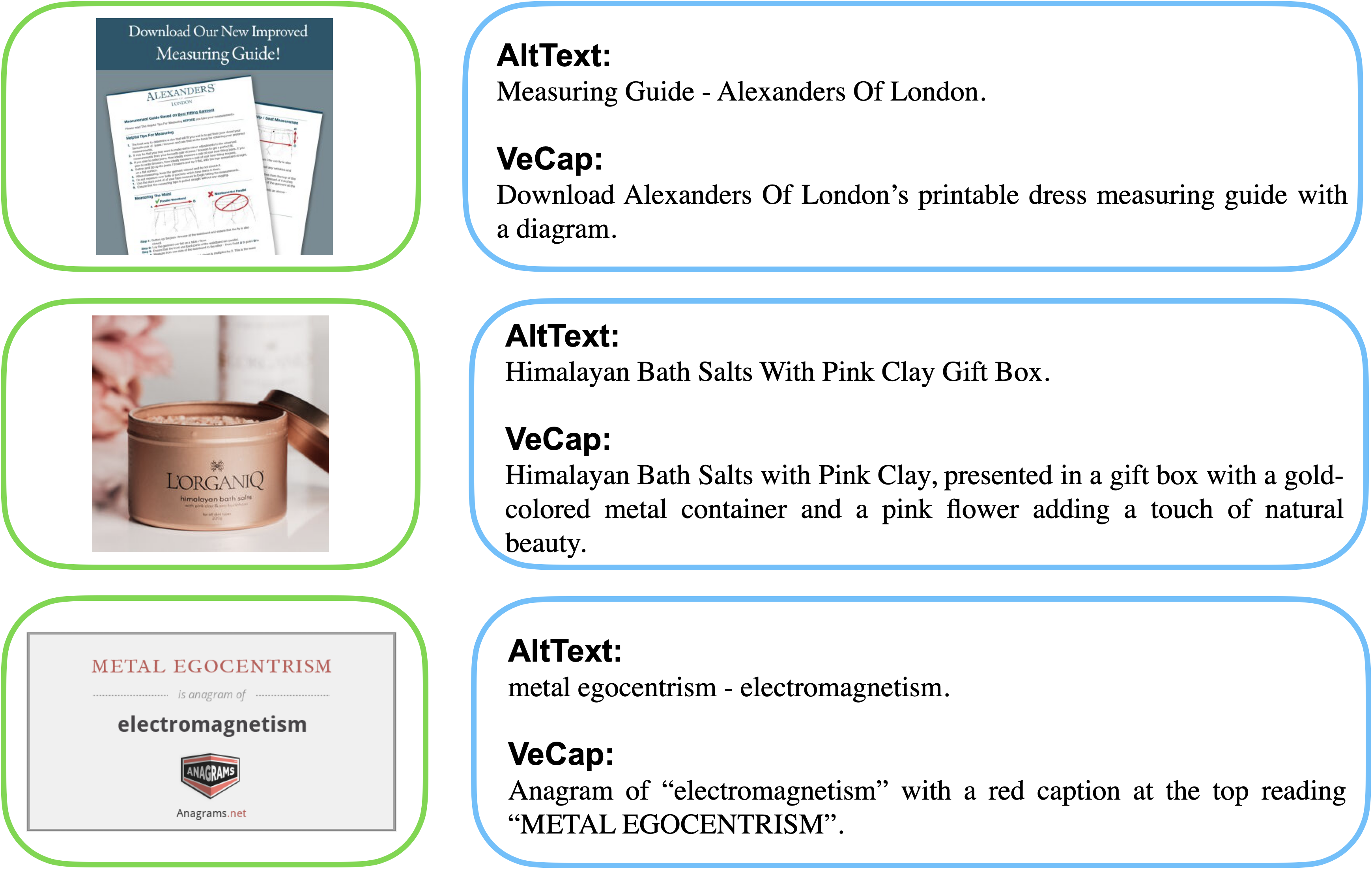} 
	
\\
\vspace{0.6cm}
\includegraphics[width=0.87\textwidth]{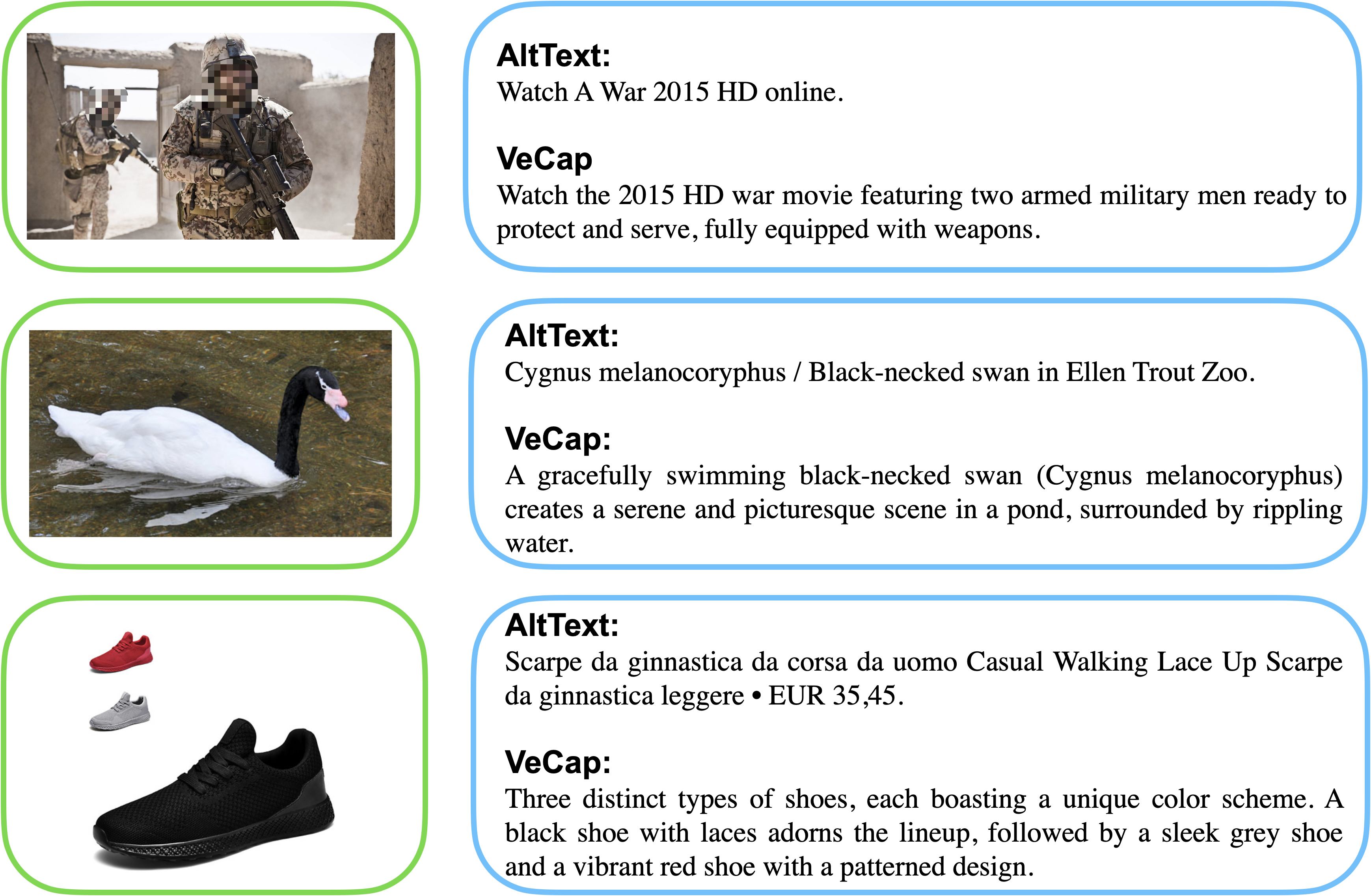} \\
	\end{tabular}
\end{center}
\vspace{-0.4cm}
\caption{More examples of VeCap captions and AltTexts.  }
\label{fig:appendix_examples}
\end{figure*}

We conduct our scalable pipeline over 200 million image-text pairs. We randomly select more examples below to show the advantages of VeCap against the original AltText in terms of visual concepts. The examples are visualized in Figure~\ref{fig:appendix_examples}.


\end{document}